\documentclass{ecai} 
\usepackage{latexsym}
\usepackage{amssymb}
\usepackage{amsmath}
\usepackage{amsthm}
\usepackage{booktabs}
\usepackage{enumitem}
\usepackage{graphicx}
\usepackage{color}
\usepackage{algorithm}
\usepackage{algorithmic}
\usepackage{multirow}

\newcommand{\BibTeX}{B\kern-.05em{\sc i\kern-.025em b}\kern-.08em\TeX}

\begin{document}

%%%%%%%%%%%%%%%%%%%%%%%%%%%%%%%%%%%%%%%%%%%%%%%%%%%%%%%%%%%%%%%%%%%%%%%%

\begin{frontmatter}

%%% Use this command to specify your submission number.
%%% In doubleblind mode, it will be printed on the first page.

\paperid{9119} 

%%% Use this command to specify the title of your paper.

\title{Dual-Center Graph Clustering with Neighbor Distribution}

%%% Use this combinations of commands to specify all authors of your 
%%% paper. Use \fnms{} and \snm{} to indicate everyone's first names 
%%% and surname. This will help the publisher with indexing the 
%%% proceedings. Please use a reasonable approximation in case your 
%%% name does not neatly split into "first names" and "surname".
%%% Specifying your ORCID digital identifier is optional. 
%%% Use the \thanks{} command to indicate one or more corresponding 
%%% authors and their email address(es). If so desired, you can specify
%%% author contributions using the \footnote{} command.

% \author[A]{\fnms{Enhao}~\snm{Cheng}\orcid{....-....-....-....}\thanks{Corresponding Author. Email: somename@university.edu.}\footnote{Equal contribution.}}
\author[A]{\fnms{Enhao}~\snm{Cheng}}
\author[A]{\fnms{Shoujia}~\snm{Zhang}}
\author[A]{\fnms{Jianhua}~\snm{Yin}\thanks{Corresponding Author. Email: jhyin@sdu.edu.cn.}} 
\author[B]{\fnms{Li}~\snm{Jin}} 
\author[C]{\fnms{Liqiang}~\snm{Nie}} 

\address[A]{Shandong University}
\address[B]{Aerospace Information Research Institute, Chinese Academy of Sciences}
\address[C]{Harbin Institute of Technology (Shenzhen)}

%%% Use this environment to include an abstract of your paper.

\begin{abstract}
Graph clustering is crucial for unraveling intricate data structures, yet it presents significant challenges due to its unsupervised nature. Recently, goal-directed clustering techniques have yielded impressive results, with contrastive learning methods leveraging pseudo-label garnering considerable attention. Nonetheless, pseudo-label as a supervision signal is unreliable and existing goal-directed approaches utilize only features to construct a single-target distribution for single-center optimization, which lead to incomplete and less dependable guidance. In our work, we propose a novel Dual-Center Graph Clustering (DCGC) approach based on neighbor distribution properties, which includes representation learning with neighbor distribution and dual-center optimization. Specifically, we utilize neighbor distribution as a supervision signal to mine hard negative samples in contrastive learning, which is reliable and enhances the effectiveness of representation learning. Furthermore, neighbor distribution center is introduced alongside feature center to jointly construct a dual-target distribution for dual-center optimization. Extensive experiments and analysis demonstrate superior performance and effectiveness of our proposed method. 
% The source code of the proposed model is publicly accessible and available in the supplementary materials.
% The code of DCGC is available at the code\&data appendix.
% The source code for our model is publicly available at \url{https://anonymous.4open.science/r/DCGC-1119}.
\end{abstract}

\end{frontmatter}

%%%%%%%%%%%%%%%%%%%%%%%%%%%%%%%%%%%%%%%%%%%%%%%%%%%%%%%%%%%%%%%%%%%%%%%%

\section{Introduction}
Clustering is a fundamental technique in data analysis, essential for grouping similar data points according to specific features. Deep clustering strives to instruct neural networks in acquiring distinctive feature representations to autonomously segment data into separate clusters without manual intervention. In deep clustering, the absence of label guidance accentuates the importance and complexity of crafting refined objective functions and sophisticated architectures. These components empower the network to assimilate comprehensive and discriminative information, unveiling the intrinsic structures within the data. 

\begin{figure}[!t]
\centering
\small
\begin{minipage}{0.49\linewidth}
\centerline{\includegraphics[width=\textwidth]{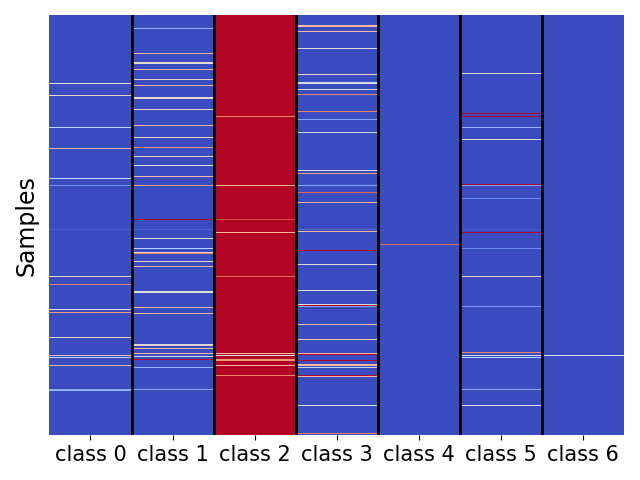}}
\centerline{(a) class 2 of CORA}
\end{minipage}
\begin{minipage}{0.49\linewidth}
\centerline{\includegraphics[width=\textwidth]{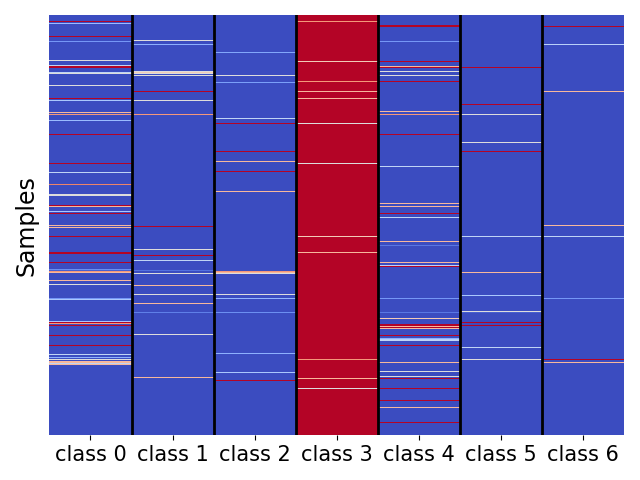}}
\centerline{(b) class 3 of CORA}
\end{minipage}
\\[1ex]  % Add some vertical space between rows
\begin{minipage}{0.49\linewidth}
\centerline{\includegraphics[width=\textwidth]{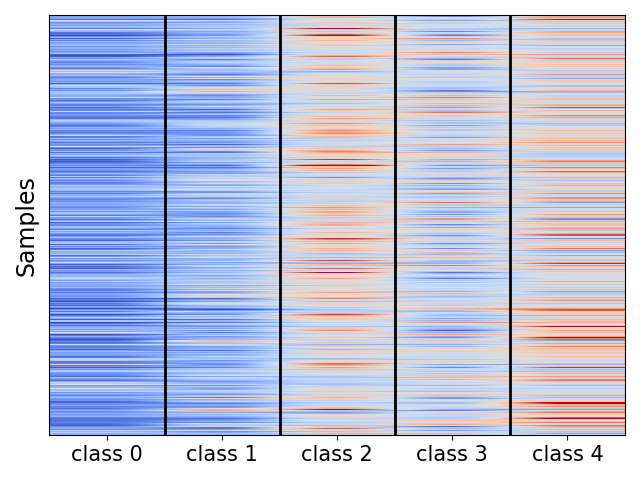}}
\centerline{(c) class 2 of SQUIRREL}
\end{minipage}
\begin{minipage}{0.49\linewidth}
\centerline{\includegraphics[width=\textwidth]{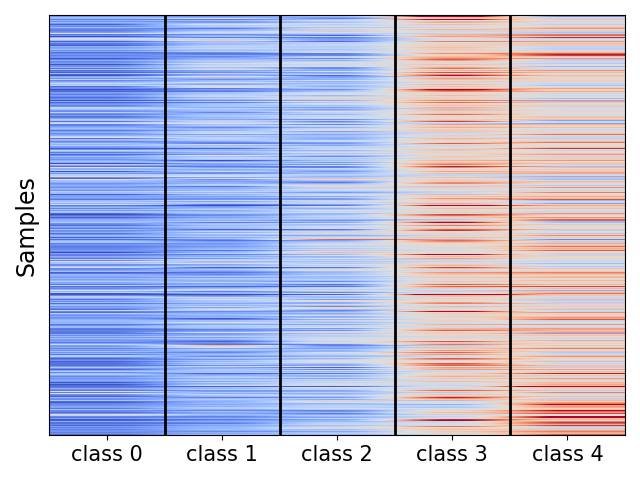}}
\centerline{(d) class 3 of SQUIRREL}
\end{minipage}
\caption{The neighbor distributions of samples from specific classes in the homophilic CORA and heterophilic SQUIRREL datasets. Each row vector represents the neighbor distribution of a node, which is defined as the class distribution aggregated from its neighbors and is expressed as a probability distribution over all classes. The intensity of the color in the visualization, along with each element of the neighbor distribution vector, indicates the proportion of neighbors belonging to each class, with values ranging from 0 to 1.}
\label{class_neighbor_distribution}
\end{figure}

Due to the complex nature of graph-structured data, tackling these challenges requires innovative approaches. Recent advancements in contrastive learning, exemplified by ProGCL~\cite{xia2021progcl} and HSAN~\cite{liu2023hard}, leverage clustering pseudo-label to enhance representation learning effectively. On the other hand, goal-directed techniques like SDCN~\cite{bo2020structural} and DFCN~\cite{tu2021deep} utilize self-supervised strategies to construct target distributions. However, pseudo-label is used as a supervision signal, and goal-directed methods rely solely on node features to construct a single-target distribution for single-center optimization, which is incomplete and lacks reliable guidance.

Given the limitations of contrastive learning and goal-directed methods, we systematically explore neighbor distribution that attract attention in graph research. Through detailed analysis and observation of various graph datasets, as illustrated in Figure \ref{class_neighbor_distribution}, we identify two key properties that stand out for their significant impact and value in clustering tasks. First, we observe intra-class consistency, where nodes within the same class tend to have highly similar neighbor distributions. For instance, in heterophilic datasets like Squirrel, nodes belonging to the same class share similar neighbor distributions, which reinforces the cohesion within clusters. This consistency provides a robust basis for grouping similar nodes in the clustering process. Second, we find that neighbor distribution is reliable and insensitive to errors in initial predictions. Thus, neighbor distributions aggregated from the class of neighbors can serve as more stable supervision signals. This reliability is especially evident in homophilic datasets like CORA, where neighbor distributions enable nodes to make accurate predictions even when their own pseudo-labels are incorrect. These two properties effectively address the challenges of clustering, forming the foundation of our proposed method.

By leveraging these insights, we utilize neighbor distribution as a supervision signal to mine hard negative samples in contrastive learning, which is reliable and enhances the effectiveness of representation learning. Furthermore, neighbor distribution center is introduced alongside feature center to jointly construct a dual-target distribution for dual-center optimization. Moreover, our method achieves superior performance across various graph structures by combining contrastive learning and goal-directed learning. Specifically, we first extract structural information through an adaptive filterbank, followed by representation learning using neighbor distribution to train the embeddings. Finally, neighbor distributions are derived from pseudo-labels, and dual-center optimization is employed for network fine-tuning. The main contributions of this paper are summarized as follows.
\begin{itemize}
    \item We propose a novel Dual-Center Graph Clustering approach with neighbor distribution (DCGC), which utilizes neighbor distribution as a supervision signal to mine hard negative samples in contrastive learning and constructs a dual-target distribution for dual-center optimization.
    \item We introduce neighbor distribution center alongside feature center to jointly construct a dual-target distribution for dual-center optimization and fine-tune representation learning, which utilizes the intra-class consistency and reliability of neighbor distribution to improve the single-target distribution. 
    \item Extensive experimental results on six datasets demonstrate the superiority and effectiveness of our model. Our method consistently outperforms state-of-the-art baselines across multiple benchmark datasets.
\end{itemize}

\section{Related Work}
\subsection{Unsupervised Graph Representation Learning}
Early methods maximize the similarity of representations among proximal nodes to learn low-dimensional node representations on graphs~\cite{perozzi2014deepwalk,grover2016node2vec,pan2018adversarially}. The GAE model~\cite{kipf2016variational} adopts a graph convolutional encoder~\cite{kipf2016semi} to learn improved representations, but the lack of label supervision limits its performance. Recent efforts have applied contrastive learning~\cite{chen2020simple} to unsupervised graph representation learning, enhancing performance by maximizing the agreement between two augmented views of the graph~\cite{zhu2020deep,you2020graph,velickovic2019deep}. For example, the AGE model~\cite{cui2020adaptive} trains the encoder by adaptively discriminating positive and negative samples. Subsequent methods, such as the graph debiased contrastive learning framework~\cite{zhao2021graph} and HSAN~\cite{liu2023hard}, have adopted novel contrastive techniques that utilize clustering results to optimize the contrastive learning network. However, these methods relying on pseudo-label lack reliability, resulting in suboptimal performance. In contrast, our approach utilizes neighbor distribution as a supervision signal to mine hard negative samples in contrastive learning, which is reliable and enhances the effectiveness of representation learning.

\begin{figure*}[!t]
\centering
\includegraphics[scale=0.49]{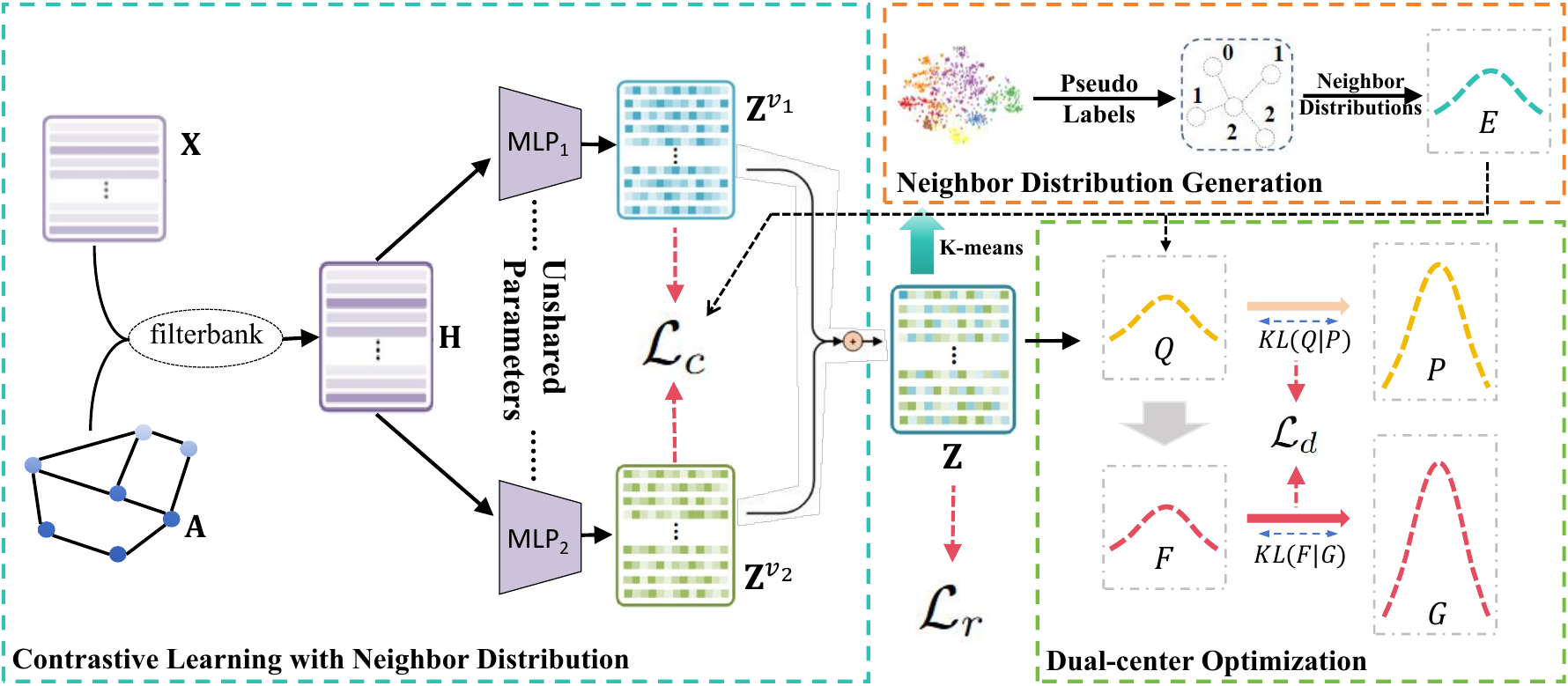}
\caption{Illustration of our proposed dual-center graph clustering approach. Contrastive learning with neighbor distribution employs an adaptive filterbank and contrastive objectives to generate node representations $\mathbf{Z}$. Then, K-means is applied to generate node neighbor distributions during neighbor distribution generation, which are utilized as supervision signals to mine hard negative samples in contrastive learning and to construct a dual-target distribution. Finally, dual-center optimization fine-tunes representation learning by aligning soft assignment with the dual-target distribution.}
\label{OVERRALL_FIGURE}
\end{figure*}

\subsection{Deep Graph Clustering}
Graph clustering has been an enduring research focus. Early deep graph clustering algorithms employ various types of autoencoders~\cite{kipf2016variational,tian2014learning,pan2019learning,ghasedi2017deep} to learn deep representations for clustering. 
Subsequently, contrastive graph clustering methods~\cite{zhao2021graph,lee2022augmentation} are introduced. Some approaches~\cite{liu2023hard,yang2023cluster} integrate clustering information~\cite{hartigan1979algorithm} to optimize feature learning, achieving promising outcomes. Another pioneering method in this area first trains a pre-trained network and then defines a target distribution to fine-tune the network with stronger guidance~\cite{xie2016unsupervised}. DAEGC focuses on attributed graphs to sufficiently explore the topological structure and node content~\cite{wang2019attributed}. Building on this, SDCN integrates autoencoder and graph convolutional network with a self-supervised mechanism to improve deep clustering performance~\cite{bo2020structural}. DFCN~\cite{tu2021deep} focuses on reliable target distribution generation and employs a triplet self-supervision strategy. However, current methods are limited by single-center optimization in the feature space, which is inherently incomplete and lacks reliable guidance. In response, our method introduces neighbor distribution center alongside feature center to jointly construct a dual-target distribution for dual-center optimization, which provides more reliable and robust clustering guidance compared to single-center methods.

\section{Methodology}
In this section, we introduce the proposed dual-center graph clustering approach with neighbor distribution. The framework is illustrated in Figure~\ref{OVERRALL_FIGURE}.

\subsection{Notations and Definitions}
Let $\mathcal{V} = \{v_1, v_2, \ldots, v_N\}$ denote a set of $N$ nodes, categorized into $C$ classes, and let $\mathcal{E}$ denote the set of edges. In matrix form, $\mathbf{X} \in \mathbb{R}^{N \times D}$ and $\mathbf{A} \in \mathbb{R}^{N \times N}$ represent the attribute matrix and the original adjacency matrix, respectively. Thus, $\mathcal{G} = \{\mathbf{X}, \mathbf{A}\}$ defines an undirected graph. The degree matrix is expressed as $\mathbf{D} = \operatorname{diag}(d_1, d_2, \ldots, d_N) \in \mathbb{R}^{N \times N}$, where $d_i = \sum_{(v_i, v_j) \in \mathcal{E}} A_{ij}$. The graph Laplacian matrix is defined as $\mathbf{L} = \mathbf{D} - \mathbf{A}$. Using the renormalization trick, the modified adjacency matrix is $\widehat{\mathbf{A}} = \mathbf{A} + \mathbf{I}$, and the symmetric normalized graph Laplacian matrix is given by $\widetilde{\mathbf{L}} = \mathbf{I} - \widehat{\mathbf{D}}^{-\frac{1}{2}} \widehat{\mathbf{A}} \widehat{\mathbf{D}}^{-\frac{1}{2}}$. 

Generally, a neural network $\mathcal{F}$ is initially trained without human annotations to embed nodes into a latent space, after which a clustering algorithm $\mathcal{C}$ is applied to partition the nodes into $K$ disjoint groups:
\begin{equation}
\Phi = \mathcal{C}(\mathcal{F}(\mathbf{A}, \mathbf{X})),
\end{equation}
where $\Phi \in \mathbb{R}^{N \times K}$ represents the cluster membership matrix for all $N$ samples. 

\subsection{Neighbor Distribution Generation}
Acknowledging the importance of intra-class consistency and reliability in neighbor distribution as highlighted in the introduction, this section explains how to calculate neighbor distribution. It is crucial to understand the subsequent application of neighbor distribution in the model. In this process, K-means~\cite{hartigan1979algorithm} is applied to the learned node embeddings to generate pseudo-labels, which are used to derive neighbor distributions. Specifically, we first generate the pseudo-labels $\mathbf{Y} \in \mathbb{R}^{N \times K}$, where $\mathbf{y}_{i}$ represents the one-hot encoding of pseudo-label for node $i$. Based on $\mathbf{Y}$, we obtain the neighbor distribution $\mathbf{e}_i$ of node $i$.
\begin{equation}
\mathbf{e}_i = \frac{1}{\left|N_i\right|} \sum_{j \in N_i}\mathbf{y}_{j}, i=1,2, \ldots, N,
\label{eq_e}
\end{equation}
where $N_i$ is the set of neighbors of node $i$. $\left|N_i\right|$ is the number of neighbors of node $i$. Based on the neighbor distribution of each node, we derive the class neighbor distribution $\boldsymbol{\pi}_k$ of cluster $k$.
\begin{equation}
\boldsymbol{\pi}_k = \frac{1}{\left|V_k\right|} \sum_{j \in V_k} \mathbf{e}_{j}, k=1,2, \ldots, K,
\label{eq_pi}
\end{equation}
where $V_k$ represents the set of nodes predicted to belong to cluster $k$. $\left|V_k\right|$ denotes the number of nodes in this set. In summary, we use the pseudo-labels of nodes to derive their neighbor distributions and the class neighbor distributions, which are utilized as supervision signals to mine hard negative samples in contrastive learning and to construct a dual-target distribution for dual-center optimization. 

\subsection{Representation Learning with Neighbor Distribution}
% In the process of attribute encoding, we follow previous work~\cite{luan2022revisiting} and filter noise in the attribute matrix $\mathbf{H}$ as formulated:
In our view, graph structures are neither purely homophilic nor heterophilic. Recent node-wise analyses indicate that different nodes may require different filters to process their neighborhood information~\cite{luan2022revisiting}. Based on this analysis, we propose the adaptive filterbank, a three-channel architecture that adaptively exploits local and node-wise information from aggregation, diversification, and identity channels. This captures low-frequency information in graph signals and preserves the shared knowledge of similar nodes.
The combined node representation $\mathbf{H}$ can be formulated as:
\begin{equation}
\begin{aligned}
\mathbf{H} &= \alpha_1 \mathbf{H}_{lp} + \alpha_2 \mathbf{H}_{hp} + \alpha_3 \mathbf{X} \\
&= \alpha_1 (\mathbf{I}-\widetilde{\mathbf{L}})^t \mathbf{X} + \alpha_2 \widetilde{\mathbf{L}}^t \mathbf{X} + \alpha_3 \mathbf{X},
\end{aligned}
\label{ACM}
\end{equation}
where $\mathbf{I} - \widetilde{\mathbf{L}}$ performs low-pass filtering~\cite{yang2021diverse} , and $\widetilde{\mathbf{L}}$ is used to perform high-pass filtering~\cite{dong2021adagnn,ma2021unified}. The last term denotes the identity channel, which preserves the original features of nodes without modification. Moreover, $t$ represents the filtering times. $\alpha_1$, $\alpha_2$, and $\alpha_3$ are learnable parameters that control the contribution of each channel.

The adaptive filterbank offers an effective framework for node representation learning in complex graph structures by adaptively adjusting these weights. Then, we design representation learning with neighbor distribution to mine hard negative samples in contrastive learning, producing better embeddings. In our model, the adaptive filterbank integrates structural information, effectively extending the graph Laplacian filter. Consequently, we employ simple multilayer perceptron (MLP) for feature augmentation.
\begin{equation}
\begin{aligned}
& \mathbf{Z}^{v_1}=\operatorname{MLP}_1(\mathbf{H}) ; \mathbf{Z}_i^{v_1}=\frac{\mathbf{Z}_i^{v_1}}{\left\|\mathbf{Z}_i^{v_1}\right\|_2}, i=1,2, \ldots, N; \\
& \mathbf{Z}^{v_2}=\operatorname{MLP}_2(\mathbf{H}) ; \mathbf{Z}_j^{v_2}=\frac{\mathbf{Z}_j^{v_2}}{\left\|\mathbf{Z}_j^{v_2}\right\|_2}, j=1,2, \ldots, N,
\end{aligned}
\label{MLP}
\end{equation}
where $\mathbf{Z}^{v_1}$ and $\mathbf{Z}^{v_2}$ denote two-view attribute embeddings of the samples and the MLP have the same architecture but unshared parameters. Previous work~\cite{gao2021simcse} has demonstrated that simple networks with unshared parameters can capture distinct semantic information. Therefore, $\mathbf{Z}^{v_1}$ and $\mathbf{Z}^{v_2}$ are utilized for feature augmentation in this process. Different the Mixup-based methods~\cite{wu2021graphmixup,yang2022mixed} which depend on carefully designed augmentation strategies, MLP-based augmentation is efficient and effective. Based on the two views $\mathbf{Z}^{v_1}$ and $\mathbf{Z}^{v_2}$, we formulate the proposed contrastive loss with neighbor distribution as follows:
\begin{equation} 
\begin{aligned}
&\mathcal{L}_c=\frac{1}{2N} \sum_{l=1}^{2} \sum_{i=1}^{N} \mathcal{L}_c(i, l) =-\log(\\
&\frac{e^{\mathcal{M}(i,i) \cdot \mathcal{S}(\mathbf{z}_i^{v_1},\mathbf{z}_i^{v_2})}}{e^{\mathcal{M}(i,i) \cdot \mathcal{S}(\mathbf{z}_i^{v_1},\mathbf{z}_i^{v_2})} +\sum_{v \in \{v_1, v_2\}, j \neq i}{e^{\mathcal{M}(i,j) \cdot \mathcal{S}(\mathbf{z}_i^{v_l},\mathbf{z}_j^v)}}}),
\end{aligned}
\label{our_loss_part}
\end{equation}
where $\mathcal{S}(, )$ denotes the cosine similarity between paired samples in the latent space. By minimizing the loss $\mathcal{L}_c$, the model pulls together identical samples from different views while pushing apart other samples.

In classical InfoNCE~\cite{oord2018representation}, all sample pairs are treated equally, which limits the network's discriminative capability. To address this issue, we propose a weighting function $\mathcal{M}$ to adjust the weights of negative sample pairs and mine hard negative samples in contrastive learning. Specifically, $\mathcal{M}$ is defined as follows:
\begin{equation} 
\mathcal{M}(i,j)=
\begin{cases}|\mathcal{K}_{i,j}-Norm(\mathcal{S}(\mathbf{z}_i,\mathbf{z}_j))|, & i \neq j; \\
1, & i = j, \end{cases}
\label{M}
\end{equation}
where $Norm$ denotes min-max normalization, and $\mathcal{K}_{i,j}$ is defined as:
\begin{equation}
\mathcal{K}_{i j}= \begin{cases}1, & \mathcal{S}(\mathbf{e}_i,\mathbf{e}_j)>\tau; \\ 0, & \text { others, }\end{cases}
\end{equation}
where $\tau$ is a hyper-parameter, and $\mathbf{e}_i$ denotes the neighbor distribution of node $i$, generated from the pseudo-label. $\mathcal{M}(i,j)$ measures the difference between neighbor distribution similarity and embedding similarity, which can up-weight the hard samples while down-weighting the easy samples. A larger difference indicates a hard sample, which should be assigned a higher weight to draw greater attention.

Additionally, we incorporate a structural reconstruction loss to further explore structural information. Specifically, we employ an inner product decoder to reconstruct the structure $\hat{A}_{ij} = \sigma(z_i^\top z_j)$, where $\sigma(\cdot)$ denotes the sigmoid function. The reconstruction loss below quantifies the error between the original matrix $A$ and the reconstructed matrix $\hat{A}$, preserving the structural information.
\begin{equation}
\begin{aligned}
\mathcal{L}_r = \sum_{i=1}^{n} \sum_{j=1}^{n} (- A_{ij} \log(\hat{A}_{ij}) - (1 - A_{ij}) \log(1 - \hat{A}_{ij})).
\end{aligned}
\label{reconstruction_loss}
\end{equation}

Finally, we combine the latent embeddings from the two views using a linear combination operation, enabling the resultant latent embeddings to be utilized for dual-center optimization.
\begin{equation}
\mathbf{Z}=\frac{1}{2}\left(\mathbf{Z}^{v_1}+\mathbf{Z}^{v_2}\right).
\label{ZZZ}
\end{equation}

\subsection{Dual-center Optimization}
To provide more reliable guidance for training the clustering network, we first utilize the embedding $\mathbf{Z}$, which integrates structural information, to generate the target distribution. The generation process consists of two steps:
\begin{align}
     \label{feature_center}
    q_{ij}&=\frac{(1+||\mathbf{z}_i-\boldsymbol{\mu}_j||^2)^{-1}}{\sum_k(1+||\mathbf{z}_i-\boldsymbol{\mu}_k||^2)^{-1}}, \\
    & \label{sharpening1}
    p_{ij}=\frac{q_{i j}^2 / \sum_i q_{i j}}{\sum_k\left(q_{i k}^2 / \sum_i q_{i k}\right)}.
\end{align}

In the first step, $\boldsymbol{\mu}_j$ represents the $j$-th pre-calculated feature center, determined using K-means. The similarity $q_{ij}$ between node $i$ and cluster $j$ in the feature space is calculated using Student’s t-distribution as kernel~\cite{van2008visualizing}, resulting in a soft assignment. The soft assignment matrix $\mathbf{Q} \in \mathbb{R}^{N \times K}$ represents the distributions of all samples, where $\mathbf{q}_i$ denotes the assignment distribution of node $i$. In the second step, to enhance the confidence of cluster assignments, we apply Eq.~\eqref{sharpening1} to drive all samples closer to their respective feature centers. Specifically, $0 \leq p_{i j} \leq 1$ is an element of the generated target distribution $\mathbf{P} \in \mathbb{R}^{N \times K}$, which indicates the probability that the $i$-th sample belongs to the $j$-th feature center.

Although current graph clustering methods can provide supervision signals~\cite{xie2016unsupervised,wang2019attributed}, the generated target distributions remains incomplete and less comprehensive. To exploit the intra-class consistency and reliability of neighbor distribution, as discussed in the introduction, we introduce neighbor distribution center alongside feature center to jointly construct a dual-target distribution for dual-center optimization.

Here, we use the soft assignment $\mathbf{q}_{i}$ as the label distribution, replacing the pseudo-label $\mathbf{y}_i$. This facilitates to derive the neighbor distribution $\mathbf{e}_i$ of node $i$ by Eq.~\eqref{eq_e}, and the class neighbor distribution $\boldsymbol{\pi}_k$ of cluster $k$ by Eq.~\eqref{eq_pi}. Then, we compute a new set of soft assignments $f_{ij}$ using neighbor distribution, which is defined as the association between the aggregated neighbor distribution $\mathbf{e}_i$ and cluster $j$ in the neighbor distribution space.
\begin{equation}
f_{ij}=\frac{(1+||\mathbf{e}_i-\boldsymbol{\pi}_j||^2)^{-1}}{\sum_k(1+||\mathbf{e}_i-\boldsymbol{\pi}_k||^2)^{-1}},
\label{neighbor_distribution_center}
\end{equation}
where $f_{ij}$ measures the similarity between neighbor distribution $\mathbf{e}_i$ and neighbor distribution center $\boldsymbol{\pi}_j$. The soft assignment matrix $\mathbf{F} \in \mathbb{R}^{N \times K}$ represents the distributions of all samples. 

The neighbor distribution center, serving as the second clustering center, is initialized using the class neighbor distribution and updated jointly with the network. In addition to driving samples closer to the feature centers in the feature space, we introduce Eq.~\eqref{sharpening2} to encourage all samples to converge toward the neighbor distribution centers in the neighbor distribution space.
\begin{equation}
g_{ij}=\frac{f_{i j}^2 / \sum_i f_{i j}}{\sum_k\left(f_{i k}^2 / \sum_i f_{i k}\right)}.
\label{sharpening2}
\end{equation}

\begin{algorithm}[t]
\small
\caption{Dual-Center Graph Clustering with Neighbor Distribution}
\textbf{Input:} Graph $G = \{ \mathbf{X}, \mathbf{A} \}$ with $N$ nodes; Number of clusters $K$; Pretraining epoch number $Epoch_1$; Epoch number $Epoch_2$; Filtering times $t$; Update interval $T$; hyper-parameters $\beta$ and $\tau$.\\
\textbf{Output}: Final clustering results $\Phi$.\\
\begin{algorithmic}[1]
\FOR{$i = 0$ to $Epoch_1 - 1$}
    \STATE Obtain the adaptive filtered attribute $\mathbf{H}$ using Eq.~\eqref{ACM}.
    \STATE Generate node embeddings $\mathbf{Z}$ by attribute encoders $\operatorname{MLP_1}$ and $\operatorname{MLP_2}$.
    \STATE Perform K-means on node embeddings $\mathbf{Z}$ to obtain pseudo-labels and neighbor distributions by Eq.~\eqref{eq_e}.
    \STATE Calculate the weighting function $\mathcal{M}$ by Eq.~\eqref{M}.
    \STATE Update the model by minimizing the combined loss $\mathcal{L}_c + \mathcal{L}_r$.
\ENDFOR
\STATE Extract the pretraining hidden embedding $\mathbf{Z}$.
\STATE Compute the feature centers $\boldsymbol{\mu}$ and the class neighbor distributions $\boldsymbol{\pi}$.
\FOR{$l = 0$ to $Epoch_2-1$}
    \STATE Calculate soft assignment distributions $\mathbf{Q}$ and $\mathbf{F}$ with $\boldsymbol{\mu}, \boldsymbol{\pi}$ according to Eq.~\eqref{feature_center} and Eq.~\eqref{neighbor_distribution_center}.
    \IF{$l \% T == 0$}
        \STATE Calculate dual-target distributions by Eq.~\eqref{sharpening1} and Eq.~\eqref{sharpening2}.
    \ENDIF
    \STATE Calculate clustering loss $\mathcal{L}_d$ according to Eq.~\eqref{L_d}.
    \STATE Update the whole framework by minimizing Eq.~\eqref{all}.
\ENDFOR
\STATE Get the clustering results $\Phi$ by applying K-means to $\mathbf{Z}$.
\STATE \textbf{return} $\Phi$
\end{algorithmic}
\label{ALGORITHM}
\end{algorithm}

We introduce a sharpening operation to encourage all samples to converge toward to the dual centers. Finally, the clustering loss compels the feature assignments $\mathbf{Q}$ and the neighbor distribution assignments $\mathbf{F}$ to align with the dual-target distributions. To achieve this, we design a dual-center loss by adapting the KL-divergence, formulated as follows:
\begin{equation}
% L_d = \lambda * KL(P || Q) + (1 - \lambda) * KL(G || F)
\mathcal{L}_d = \sum_{i=1}^{N} \sum_{j=1}^{K} (\lambda *p_{ij} \log \frac{p_{ij}}{q_{ij}} + (1 - \lambda) * g_{ij} \log \frac{g_{ij}}{f_{ij}}),
\label{L_d}
\end{equation}
where $N$ is the number of nodes and $K$ is the number of clusters. $\lambda$ is a learnable weight parameter that mitigates noise from conflicting information and balances the contributions of the dual centers to the loss. Since the dual-target distribution is generated without human guidance, the dual-center approach iteratively optimizes and updates the node features.

\begin{table*}[h]
    \centering
    \caption{Statistics information of datasets.}
    \label{tab:dataset_stats}
    \begin{tabular}{l r r r r r}
        \toprule
        \textbf{Graph datasets} & \textbf{Nodes} & \textbf{Dims.} & \textbf{Edges} & \textbf{Clusters} & \textbf{Homophily Ratio} $(h)$ \\
        \midrule
        \multicolumn{5}{l}{\textbf{High Homophilic Graphs} $(h\geq0.7)$} \\
        Cora      & 2708  & 1433 & 5429   & 7 & 0.8137 \\
        Cite  & 3327  & 3703 & 4732   & 6 & 0.7392 \\
        % ACM       & 3025  & 1870 & 29281  & 3 & 0.8207 \\
        AMAP      & 7650  & 745  & 119081 & 8 & 0.8272 \\
        UAT       & 1190   & 239  & 13599   & 4 & 0.7105 \\
        \midrule
        \multicolumn{5}{l}{\textbf{Low-Moderate Homophilic Graphs} $(0.3<h<0.7)$} \\
        EAT       & 399   & 203  & 5994   & 4 & 0.4238 \\
        BAT       & 131   & 81  & 1038   & 4 & 0.4656 \\
        \midrule
        \multicolumn{5}{l}{\textbf{Heterophilic Graphs} $(h\leq0.3)$} \\
        Texas      & 183  & 1703 & 325   & 5 & 0.0614 \\
        Cornell    & 183  & 1703 & 298   & 5 & 0.1220 \\
        Wisconsin  & 251  & 1703 & 515   & 5 & 0.1703 \\
        % Washington & 230  & 1703 & 786   & 5 & 0.1434 \\
        % Twitch     & 1912 & 2545 & 31299 & 2 & 0.5660 \\
        Squirrel   & 5201 & 2089 & 217073 & 5 & 0.2234 \\
        \bottomrule
    \end{tabular}
\end{table*}

\begin{table*}[!t]
\centering
\caption{The average clustering performance of ten runs on six homophilic datasets. The performance is evaluated by four metrics with mean value and standard deviation. The bold and underlined values indicate the best and runner-up results, respectively.}
\resizebox{\textwidth}{!}{
% \fontsize{9}{11}\selectfont
% \setlength{\tabcolsep}{1mm}
% \scalebox{0.9}{
\begin{tabular}{c|c|cccc|cccccc|c}
\toprule
\multirow{2}{*}{Dataset} & \multirow{2}{*}{Metric} & \multicolumn{4}{c|}{Goal-directed Graph Clustering} & \multicolumn{6}{c|}{Contrastive Deep Graph Clustering} \\
 & & DAEGC & ARGA & SDCN & DFCN & AGE & AGC-DRR & ProGCL & NCLA & GraphLearner & CCGC & DCGC \\
\midrule
        \multirow{4}{*}{CORA} & ACC         &70.43$\pm$0.36&71.04$\pm$0.25&35.60$\pm$2.83&36.33$\pm$0.49&\underline{73.50$\pm$1.83}&40.62$\pm$0.55&57.13$\pm$1.23&51.09$\pm$1.25&72.51$\pm$1.05&73.14$\pm$1.18&\textbf{78.68$\pm$0.80}\\
                              & NMI &52.89$\pm$0.69&51.06$\pm$0.52&14.28$\pm$1.91&19.36$\pm$0.87&\underline{57.58$\pm$1.42}&18.74$\pm$0.73&41.02$\pm$1.34&31.80$\pm$0.78&57.36$\pm$1.05&55.44$\pm$1.80&\textbf{60.36$\pm$0.70}\\
                              & ARI &49.63$\pm$0.43&47.71$\pm$0.33&07.78$\pm$3.24&04.67$\pm$2.10&\underline{50.10$\pm$2.14}&14.80$\pm$1.64&30.71$\pm$2.70&36.66$\pm$1.65&49.96$\pm$1.05&49.70$\pm$2.26&\textbf{58.97$\pm$1.59}\\
                              & F1  &68.27$\pm$0.57&69.27$\pm$0.39&24.37$\pm$1.04&26.16$\pm$0.50&69.28$\pm$1.59&31.23$\pm$0.57&45.68$\pm$1.29&51.12$\pm$1.12&\underline{71.27$\pm$1.17}&70.13$\pm$3.06&\textbf{77.15$\pm$0.59}\\
        \midrule
        \multirow{4}{*}{CITE} & ACC &64.54$\pm$1.39&61.07$\pm$0.49&65.96$\pm$0.31&69.50$\pm$0.20&69.43$\pm$0.24&68.32$\pm$1.83&65.92$\pm$0.80&59.23$\pm$2.32&65.86$\pm$1.55&\underline{70.40$\pm$0.85}&\textbf{72.38$\pm$0.54}\\
                              & NMI &36.41$\pm$0.86&34.40$\pm$0.71&38.71$\pm$0.32&43.90$\pm$0.20&\underline{44.93$\pm$0.53}&43.28$\pm$1.41&39.59$\pm$0.39&36.68$\pm$0.89&40.78$\pm$1.11&44.02$\pm$0.64&\textbf{46.37$\pm$0.48}\\
                              & ARI &37.78$\pm$1.24&40.17$\pm$0.43&40.17$\pm$0.43&45.31$\pm$0.41&34.18$\pm$1.73&\underline{47.64$\pm$0.30}&36.16$\pm$1.11&33.37$\pm$0.53&39.70$\pm$1.67&44.91$\pm$1.77&\textbf{48.51$\pm$0.73}\\
                              & F1  &62.20$\pm$1.32&58.23$\pm$0.31&63.62$\pm$0.24&64.30$\pm$0.20&\underline{64.45$\pm$0.27}&\textbf{64.82$\pm$1.60}&57.89$\pm$1.98&52.67$\pm$0.64&62.01$\pm$1.04&61.33$\pm$0.68&63.92$\pm$1.70\\
        \midrule
        \multirow{4}{*}{AMAP} & ACC &75.96$\pm$0.23&69.28$\pm$2.30&53.44$\pm$0.81&76.82$\pm$0.23&75.98$\pm$0.68&76.81$\pm$1.45&51.53$\pm$0.38&67.18$\pm$0.75&77.24$\pm$0.87&\underline{77.39$\pm$0.34}&\textbf{79.90$\pm$0.97}\\
                              & NMI &65.25$\pm$0.45&58.36$\pm$2.76&44.85$\pm$0.83&66.23$\pm$1.21&65.38$\pm$0.61&66.54$\pm$1.24&39.56$\pm$0.39&63.63$\pm$1.07&67.12$\pm$0.92&\underline{67.53$\pm$0.54}&\textbf{68.17$\pm$1.34}\\
                              & ARI &58.12$\pm$0.24&44.18$\pm$4.41&31.21$\pm$1.23&58.28$\pm$0.74&55.89$\pm$1.34&\underline{60.15$\pm$1.56}&34.18$\pm$0.89&46.30$\pm$1.59&58.14$\pm$0.82&58.20$\pm$0.58&\textbf{62.46$\pm$1.83}\\
                              & F1  &69.87$\pm$0.54&64.30$\pm$1.95&50.66$\pm$1.49&71.25$\pm$0.31&71.74$\pm$0.93&71.03$\pm$0.64&31.97$\pm$0.44&\textbf{73.04$\pm$1.08}&73.02$\pm$2.34&71.84$\pm$0.43&\underline{73.03$\pm$1.09}\\
        \midrule
        \multirow{4}{*}{BAT}  & ACC &52.67$\pm$0.00&67.86$\pm$0.80&53.05$\pm$4.63&55.73$\pm$0.06&56.68$\pm$0.76&47.79$\pm$0.02&55.73$\pm$0.79&47.48$\pm$0.64&64.20$\pm$0.53&\underline{69.08$\pm$1.94}&\textbf{81.15$\pm$0.35}\\
                              & NMI &21.43$\pm$0.35&\underline{49.09$\pm$0.54}&25.74$\pm$5.71&48.77$\pm$0.51&36.04$\pm$1.54&19.91$\pm$0.24&28.69$\pm$0.92&24.36$\pm$1.54&42.14$\pm$0.91&47.76$\pm$2.39&\textbf{57.84$\pm$0.21}\\
                              & ARI &18.18$\pm$0.29&\underline{42.02$\pm$1.21}&21.04$\pm$4.97&37.76$\pm$0.23&26.59$\pm$1.83&14.59$\pm$0.13&21.84$\pm$1.34&24.14$\pm$0.98&35.43$\pm$1.23&40.43$\pm$1.74&\textbf{57.19$\pm$0.56}\\
                              & F1  &52.23$\pm$0.03&67.02$\pm$1.15&46.45$\pm$5.90&50.90$\pm$0.12&55.07$\pm$0.80&42.33$\pm$0.51&56.08$\pm$0.89&42.25$\pm$0.34&61.30$\pm$0.88&\underline{67.83$\pm$3.42}&\textbf{81.04$\pm$0.31}\\
        \midrule
        \multirow{4}{*}{EAT}  & ACC &36.89$\pm$0.15&52.13$\pm$0.00&39.07$\pm$1.51&49.37$\pm$0.19&47.26$\pm$0.32&37.37$\pm$0.11&43.36$\pm$0.87&36.06$\pm$1.24&47.47$\pm$3.11&\underline{53.96$\pm$0.71}&\textbf{57.89$\pm$0.27}\\
                              & NMI &05.57$\pm$0.06&22.48$\pm$1.21&08.83$\pm$2.54&\underline{32.90$\pm$0.41}&23.74$\pm$0.90&07.00$\pm$0.85&23.93$\pm$0.45&21.46$\pm$1.80&23.98$\pm$2.71&31.17$\pm$0.52&\textbf{35.26$\pm$0.25}\\
                              & ARI &05.03$\pm$0.08&17.29$\pm$0.50&06.31$\pm$1.95&23.25$\pm$0.18&16.57$\pm$0.46&04.88$\pm$0.91&15.03$\pm$0.98&21.48$\pm$0.64&20.31$\pm$1.69&\underline{25.36$\pm$0.43}&\textbf{28.19$\pm$0.27}\\
                              & F1  &34.72$\pm$0.16&\underline{52.75$\pm$0.07}&33.42$\pm$3.10&42.95$\pm$0.04&45.54$\pm$0.40&35.20$\pm$0.17&42.54$\pm$0.45&31.25$\pm$0.96&43.80$\pm$0.67&51.02$\pm$2.12&\textbf{58.34$\pm$0.24}\\
        \midrule
        \multirow{4}{*}{UAT}  & ACC &52.29$\pm$0.49&49.31$\pm$0.15&52.25$\pm$1.91&33.61$\pm$0.09&\underline{52.37$\pm$0.42}&42.64$\pm$0.31&45.38$\pm$0.58&45.38$\pm$1.15&48.25$\pm$1.34&46.06$\pm$0.53&\textbf{58.31$\pm$0.74}\\
                              & NMI &21.33$\pm$0.44&25.44$\pm$0.31&21.61$\pm$1.26&\underline{26.49$\pm$0.41}&23.64$\pm$0.66&11.15$\pm$0.24&22.04$\pm$2.23&24.49$\pm$0.57&20.29$\pm$3.35&15.39$\pm$1.26&\textbf{30.13$\pm$0.96}\\
                              & ARI &20.50$\pm$0.51&16.57$\pm$0.31&\underline{21.63$\pm$1.49}&11.87$\pm$0.23&20.39$\pm$0.70&09.50$\pm$0.25&14.74$\pm$1.99&21.34$\pm$0.78&17.93$\pm$3.47&9.06$\pm$0.50&\textbf{25.38$\pm$1.03}\\
                              & F1  &\underline{50.33$\pm$0.64}&50.26$\pm$0.16&45.59$\pm$3.54&25.79$\pm$0.29&50.15$\pm$0.73&35.18$\pm$0.32&39.30$\pm$1.82&30.56$\pm$0.25&47.05$\pm$2.44&42.13$\pm$0.61&\textbf{58.63$\pm$0.66}\\
        \bottomrule
\end{tabular}
}
\label{Experimental Result}
\end{table*}

\begin{figure*}[!t]
\footnotesize
\begin{minipage}{0.16\linewidth}
\centerline{\includegraphics[width=\textwidth]{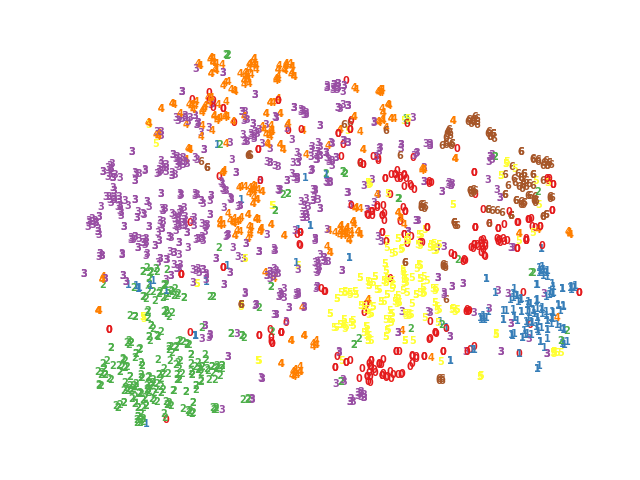}}
\centerline{(a) DAEGC}
\end{minipage}
\begin{minipage}{0.16\linewidth}
\centerline{\includegraphics[width=\textwidth]{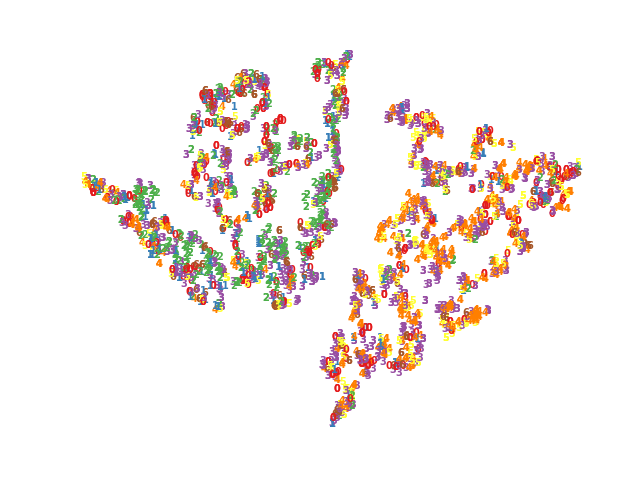}}
\centerline{(b) SDCN}
\end{minipage}
\begin{minipage}{0.16\linewidth}
\centerline{\includegraphics[width=\textwidth]{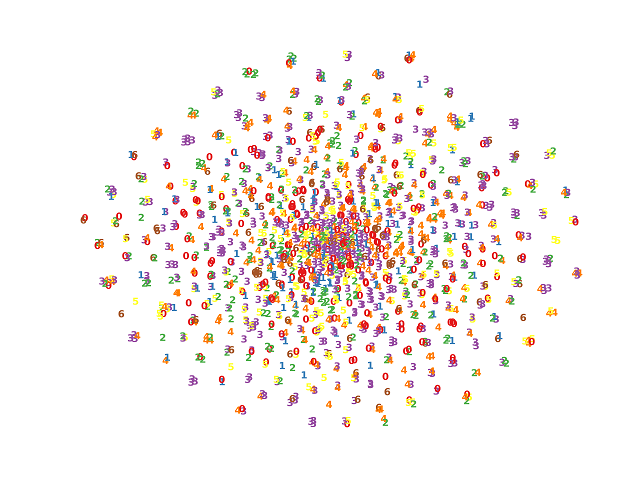}}
\centerline{(c) AFGRL}
\end{minipage}
\begin{minipage}{0.16\linewidth}
\centerline{\includegraphics[width=\textwidth]{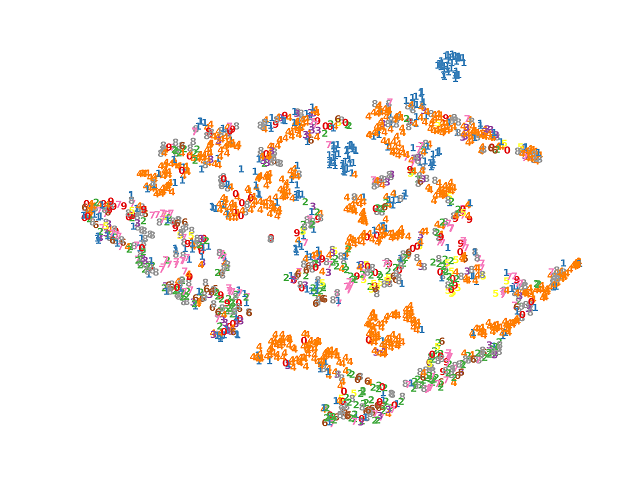}}
\centerline{(d) GDCL}
\end{minipage}
\begin{minipage}{0.16\linewidth}
\centerline{\includegraphics[width=\textwidth]{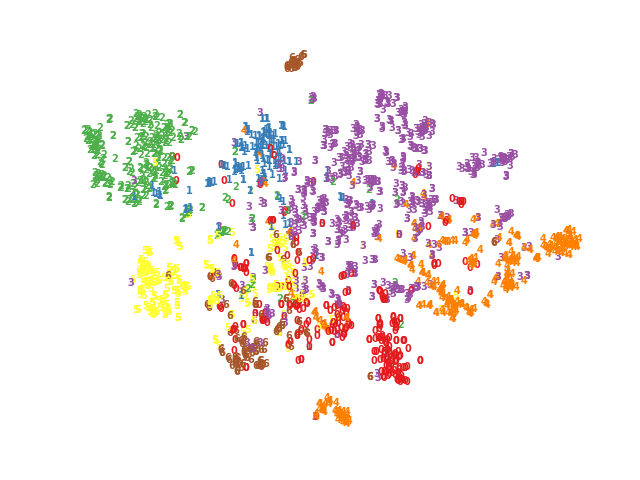}}
\centerline{(e) ProGCL}
\end{minipage}
\begin{minipage}{0.16\linewidth}
\centerline{\includegraphics[width=\textwidth]{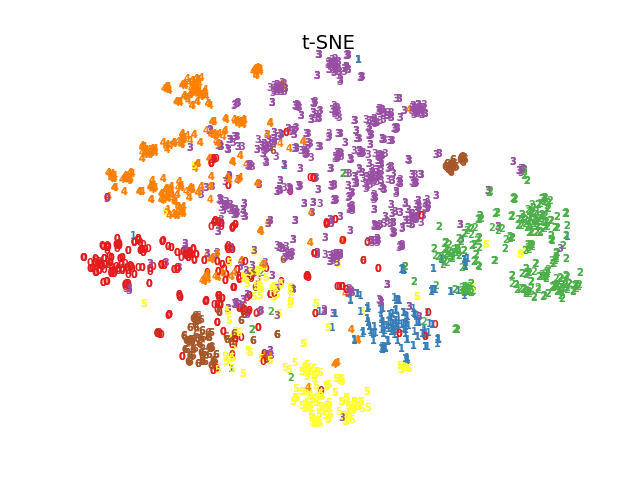}}
\centerline{(g) Ours}
\end{minipage}
\caption{The 2D $t$-SNE visualization of six methods on the CORA dataset.}
\label{t_SNE}  
\end{figure*}

\begin{table*}[htbp]
\centering
\caption{The average clustering performance of ten runs on four heterophilic datasets. }
% \fontsize{9}{11}\selectfont
% \setlength{\tabcolsep}{1mm}
% \scalebox{0.9}{
\begin{tabular}{c|c|ccccc|c}
\toprule
\textbf{Dataset} & \textbf{Metric} & DAEGC & DFCN & AGC-DRR & HSAN & CCGC & DCGC \\
\midrule
\multirow{2}{*}{Wisconsin}& ACC & 39.40$\pm$0.38 & \underline{49.84$\pm$0.12} & 49.20$\pm$3.23 & 47.93$\pm$0.40 & 45.82$\pm$0.12 & \textbf{64.54$\pm$1.75} \\
                          & NMI & \underline{14.69$\pm$0.39} & 9.17$\pm$0.06 &  8.18$\pm$4.33 & 9.93$\pm$0.80 & 7.61$\pm$0.13 & \textbf{37.13$\pm$1.86} \\
\midrule
\multirow{2}{*}{Squirrel} & ACC & 23.76$\pm$0.06 & 25.12$\pm$0.30 & 24.60$\pm$0.79 & 25.97$\pm$0.49 & \underline{30.42$\pm$0.13} &\textbf{33.57$\pm$0.05}\\
                          & NMI & 0.65$\pm$0.08 & 1.73$\pm$0.11 &  0.97$\pm$0.27 & 2.39$\pm$0.39  & \underline{5.19$\pm$0.11} & \textbf{7.96$\pm$0.48}\\
\midrule
\multirow{2}{*}{Cornell}  & ACC & 37.27$\pm$3.56 & 47.54$\pm$0.00 & \textbf{55.63$\pm$1.29} & 45.52$\pm$0.49 & 44.59$\pm$2.53 & \underline{45.90$\pm$1.89}\\
                          & NMI & \underline{13.69$\pm$2.38} & 3.03$\pm$0.00 &  5.50$\pm$2.50 & 3.78$\pm$0.33 & 4.27$\pm$0.65 & \textbf{14.14$\pm$2.81}\\
\midrule
\multirow{2}{*}{Texas}    & ACC & 32.24$\pm$2.53 & 58.42$\pm$2.30 & 56.12$\pm$1.91 & \underline{59.67$\pm$0.33} & 48.52$\pm$0.41 & \textbf{67.94$\pm$2.01}\\
                          & NMI & 6.75$\pm$2.24 & \underline{15.49$\pm$0.92} &  6.89$\pm$4.42 & 14.49$\pm$0.83 & 13.13$\pm$1.00 & \textbf{38.69$\pm$2.14}\\
\bottomrule
\end{tabular}
\label{Heterophily_Experimental_Result}
\end{table*}

\subsection{Joint Embedding and Clustering Optimization}
We jointly optimize representation learning and clustering by defining the total objective function as:
\begin{equation}
\mathcal{L} = \beta \mathcal{L}_c + (1 - \beta) \mathcal{L}_r + \gamma \mathcal{L}_d,
\label{all}
\end{equation}
where $\mathcal{L}_c$ represents the contrastive loss, $\mathcal{L}_r$ denotes the reconstruction loss, and $\mathcal{L}_d$ is the dual-center loss. The coefficients $\beta \in [0,\ 1]$ and $\gamma \geq 0$ control the balance among these terms. The proposed method is summarized in Algorithm~\ref{ALGORITHM}.

\section{Experiments}
\subsection{Summary of Datasets}
To evaluate the effectiveness of our proposed DCGC, we conduct experiments on ten benchmark datasets, including high homophilic datasets: Citation Network (CORA, CITE)~\cite{sen2008collective}, Amazon Photo (AMAP)~\cite{shchur2018pitfalls}, and USA Air Traffic (UAT); low-moderate homophilic datasets: Brazil Air Traffic (BAT) and Europe Air Traffic (EAT)~\cite{ribeiro2017struc2vec}; and heterophilic datasets: Texas, Cornell, Wisconsin~\cite{pei2020geom}, and Squirrel~\cite{rozemberczki2021multi}. The use of real-world datasets with varying levels of homophily enables a more comprehensive evaluation of the effectiveness of our model. The statistical summaries of these datasets are provided in Table~\ref{tab:dataset_stats}.

% \begin{itemize}
%     \item \textbf{Cora and CITE} \cite{sen2008collective} are classic academic citation networks characterized by high homophily, where nodes represent papers and edges represent citation relationships.
%     \item \textbf{Amazon Photo (AMAP)} \cite{shchur2018pitfalls} is derived from the Amazon co-purchase graph, where nodes represent products, edges indicate whether two products are frequently co-purchased, features are product reviews encoded using a bag-of-words model, and labels correspond to predefined product classes.
%     \item \textbf{Brazil AirTraffic (BAT), Europe Air-Traffic (EAT)}, and \textbf{USA AirTraffic (UAT)} \cite{ribeiro2017struc2vec} represent the air traffic networks of Brazil, Europe, and the United States, respectively. Here, nodes represent airports and edges reflect flight connections.
% \end{itemize}

\subsection{Experimental Setup}
The method is implemented using the PyTorch framework and runs on an NVIDIA A100 Tensor Core GPU. The total number of training epochs is set to 400, and ten independent runs are conducted for all methods. For the baselines, we use the original settings from their source code to reproduce the results. For the proposed model, the pretraining phase consists of 300 epochs, followed by 100 epochs of dual-center optimization. The attribute encoders consist of two one-layer MLPs with unshared parameters, each containing 500 hidden units for the UAT/AMAP datasets and 1500 for the other datasets, as described in~\cite{liu2023hard}. To ensure stability in dual-center optimization, the dual-target distributions are updated every five iterations during our experiments. Clustering performance is evaluated using four widely used metrics: ACC, NMI, ARI, and F1~\cite{liliang_2,ZHOU_1}. The code of DCGC is available at the code\&data appendix.

\subsection{Performance Comparison}
To demonstrate the superiority of DCGC, we conduct extensive experiments on ten benchmark datasets. Specifically, we categorize nine state-of-the-art deep graph clustering methods into two types: contrastive deep graph clustering and goal-directed graph clustering.
AGE~\cite{cui2020adaptive} and AGC-DRR~\cite{gong2022attributed} are classic contrastive deep graph clustering methods. Additionally, studies have explored sample selection strategies in contrastive learning to enhance its effectiveness~\cite{xia2021progcl,shen2023neighbor,yang2024graphlearner,liu2023hard,yang2023cluster}. DAEGC~\cite{wang2019attributed} and ARGA~\cite{pan2019learning} are representative methods in goal-directed graph clustering. SDCN~\cite{bo2020structural} and DFCN~\cite{tu2021deep} focus on improving the target distribution. In the proposed method, we leverage neighbor distribution as a supervisory signal for representation learning and introduce the neighbor distribution center alongside the feature center. This dual-center approach enhances the target distribution and effectively reduces bias.

Based on the results in Table~\ref{Experimental Result} and Table~\ref{Heterophily_Experimental_Result}, we draw the following conclusions: 
1) DCGC demonstrates superior performance compared to goal-directed graph clustering methods. Specifically, previous approaches rely exclusively on the feature center, neglecting the integration of structural information during optimization, which results in sub-optimal performance. In contrast, DCGC leverages the intra-class consistency and reliability of neighbor distribution through dual-center optimization. The dual centers, consisting of the feature center and the neighbor distribution center, successfully enhance the ability of network to capture structural information and significantly improve clustering performance. 2) It is obvious that contrastive deep graph methods are not comparable to ours. Contrastive methods typically utilize representations learned through a contrastive objective, followed by classical clustering techniques (e.g., K-means) to generate clustering results. However, there is an obvious gap between contrastive learning and the clustering task, which results in sub-optimal performance. In contrast, DCGC employs the dual-center objective to fine-tune the process of contrastive learning, effectively bridging the gap between contrastive learning and the clustering task. As a result, the proposed model significantly improves the clustering performance of the existing contrastive methods. 3) Furthermore, DCGC outperforms existing methods across all metrics, primarily due to the introduction of neighbor distribution. Specifically, we leverage neighbor distribution as a supervisory signal to mine hard negative samples in contrastive learning and construct a dual-target distribution for dual-center optimization. The performance improvement is particularly significant on heterophilic datasets. For instance, on the Wisconsin dataset, our approach achieves a 15\% improvement in ACC and a 23\% improvement in NMI compared to baseline methods. In summary, these experimental results validate the effectiveness of the proposed method, highlighting its significance in advancing graph clustering techniques.

\begin{figure}[!t]
\centering
\small
\begin{minipage}{0.49\linewidth}
% \centerline{Positive Sample Pairs}
\centerline{\includegraphics[width=0.95\textwidth]{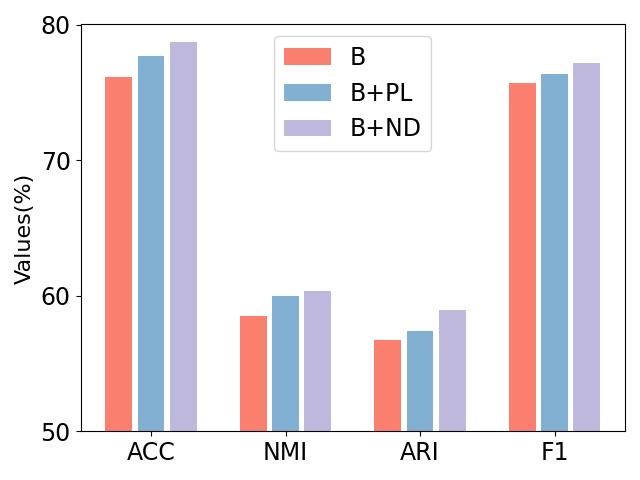}}
\centerline{(a) CORA}
\centerline{\includegraphics[width=0.95\textwidth]{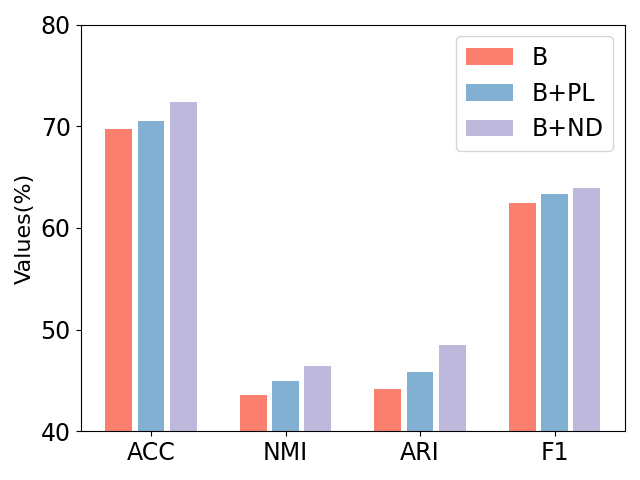}}
\centerline{(c) CITE}
\end{minipage}
\begin{minipage}{0.49\linewidth}
% \centerline{Negative Sample Pairs}
\centerline{\includegraphics[width=0.95\textwidth]{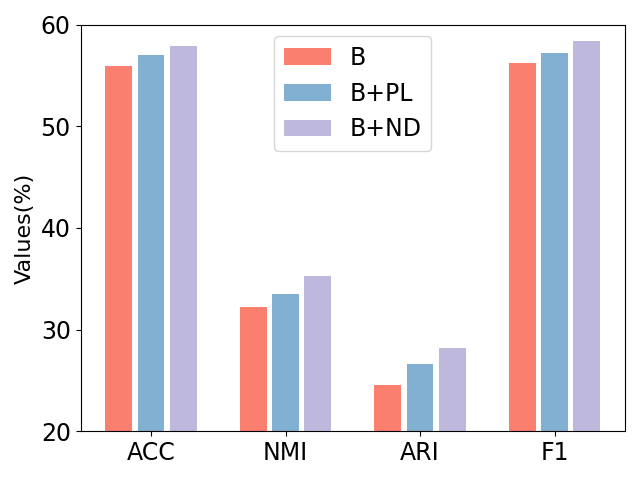}}
\centerline{(b) EAT}
\centerline{\includegraphics[width=0.95\textwidth]{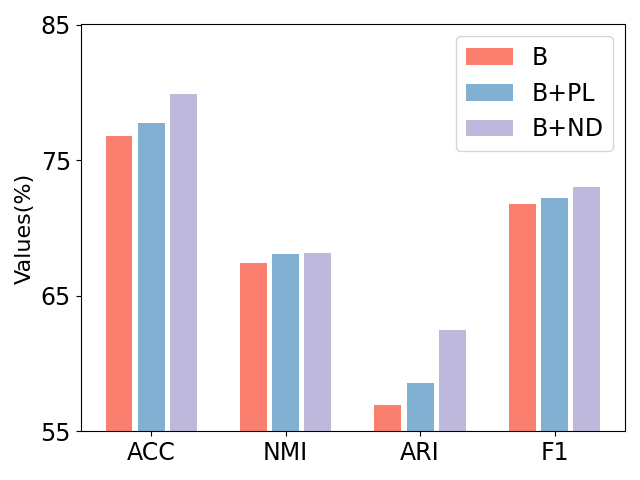}}
\centerline{(d) AMAP}
\end{minipage}
\caption{Ablation studies of the proposed representation learning with neighbor distribution on four datasets.}
\label{ablation_bar}
\end{figure}

\subsection{Ablation Study}
In this section, we perform ablation studies to evaluate the effectiveness of DCGC. As shown in Figure~\ref{ablation_bar}, ``B'' denotes the DCGC model utilizing the original representation learning network without any supervision signals. ``B+PL'' and ``B+ND'' indicate the incorporation of pseudo-label and neighbor distribution, respectively, as supervision signals in representation learning. The results show that the model utilizing neighbor distribution outperforms the other methods across all metrics. This validates the effectiveness and reliability of incorporating neighbor distribution into representation learning. Neighbor distribution serve as a supervision signal, significantly enhancing the process of representation learning.

\begin{figure}[!t]
\centering
\small
\begin{minipage}{0.49\linewidth}
\centerline{\includegraphics[width=0.98\textwidth]{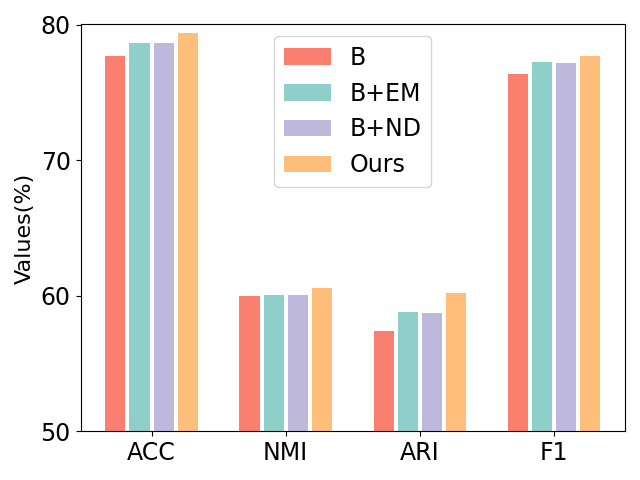}}
\centerline{(a) CORA}
\end{minipage}
\begin{minipage}{0.49\linewidth}
\centerline{\includegraphics[width=0.98\textwidth]{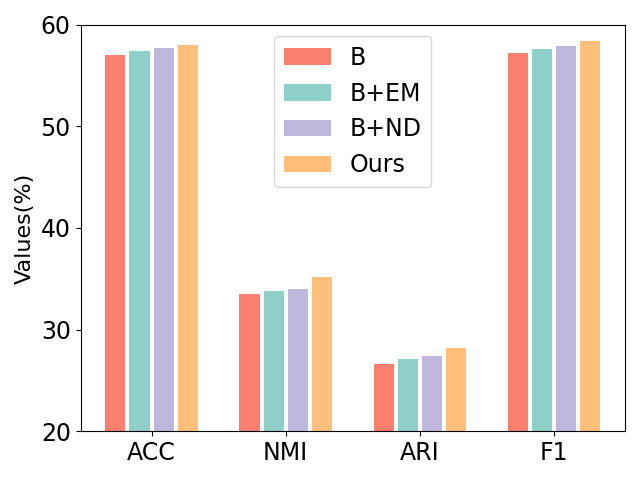}}
\centerline{(b) EAT}
\end{minipage}
\caption{Ablation studies of the proposed dual-center optimization.}
\label{ablation_plot}
\end{figure}

In Figure~\ref{ablation_plot}, ``B'' represents the baseline method without goal-directed optimization. ``B+ED'' refers to the optimization method using a single feature center, while ``B+ND'' represents the optimization method using a single neighbor distribution center. ``Ours'' denotes DCGC, which employs dual-center optimization. The results reveal the following: 1) Compared to representation learning without goal-directed optimization, single-center optimization, such as using the feature center or the neighbor distribution center, improves clustering performance across all metrics. ACC and F1 scores exhibit an approximate improvement of 1\% on the CORA dataset, with similar performances observed at the feature and neighbor distribution centers.  2) When comparing the proposed dual-center optimization with single-center optimization, the dual-center approach demonstrates improve performance across all datasets. This improvement is attributed to the complementarity of the dual-target distributions, resulting in superior performance metrics. For instance, on the EAT dataset, dual-center optimization enhances accuracy and NMI by approximately 1\%. These results validate the effectiveness and competitiveness of dual-center optimization.

\begin{figure}[!t]
\centering
\small
\begin{minipage}{0.49\linewidth}
\centerline{\includegraphics[width=0.98\textwidth]{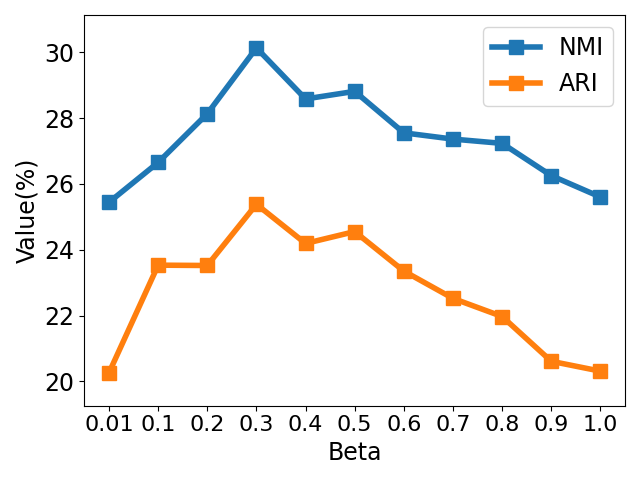}}
\centerline{(a) UAT}
\end{minipage}
\begin{minipage}{0.49\linewidth}
\centerline{\includegraphics[width=0.98\textwidth]{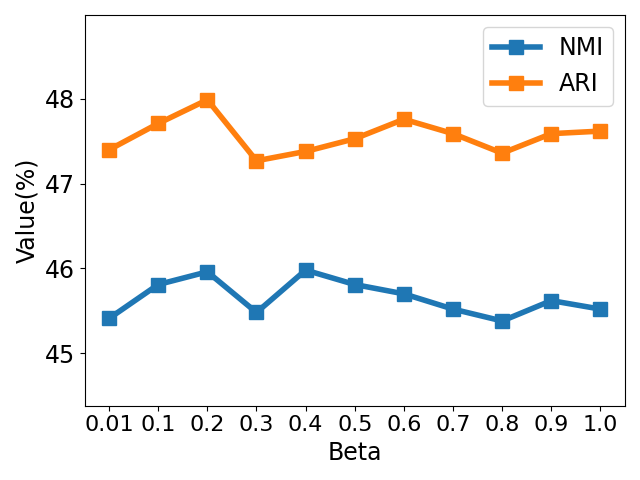}}
\centerline{(b) CITE}
\end{minipage}
\caption{Analysis of the hyper-parameter $\beta$.}
\label{beta_an}
\end{figure}

\subsection{Hyper-parameter Analysis}
\begin{figure}[!t]
\centering
\small
\begin{minipage}{0.49\linewidth}
\centerline{\includegraphics[width=0.98\textwidth]{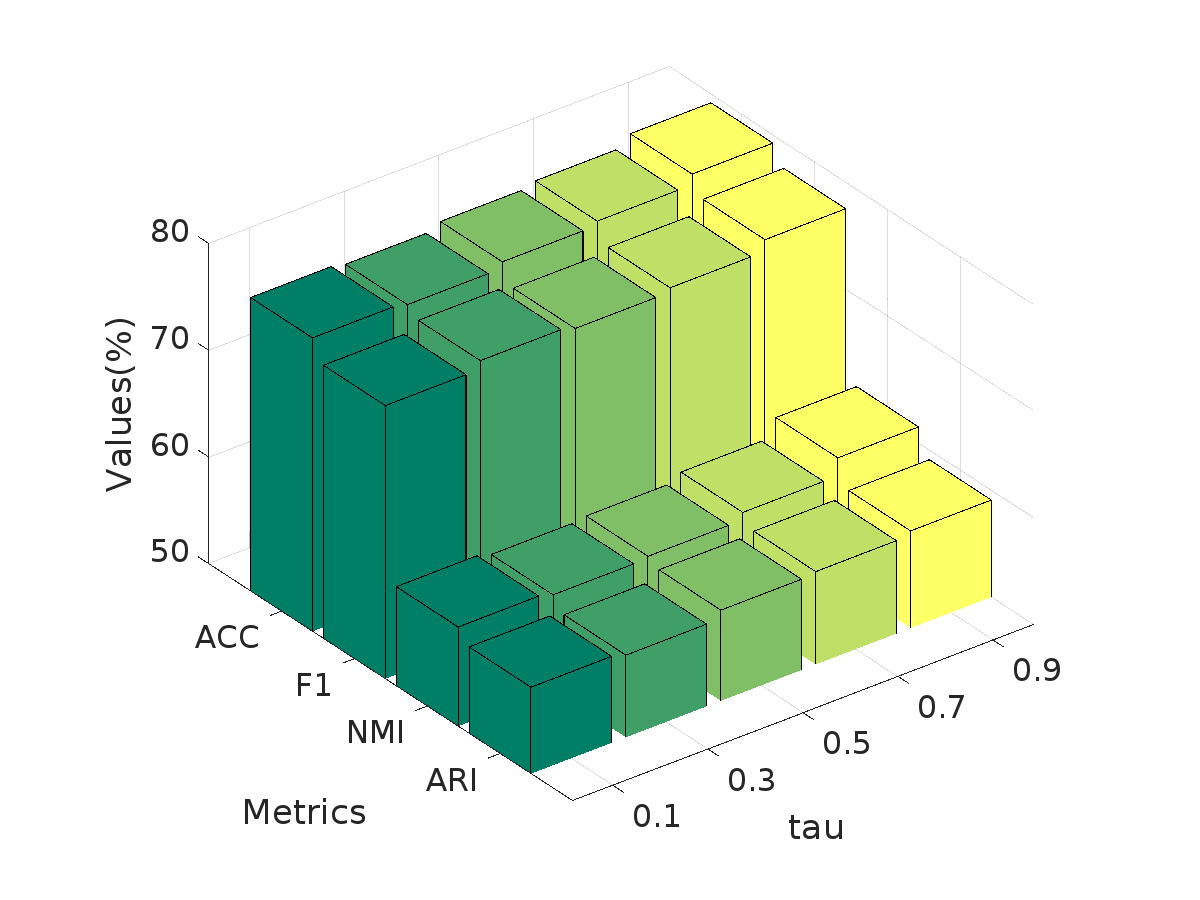}}
\centerline{(a) CORA}
\centerline{\includegraphics[width=0.98\textwidth]{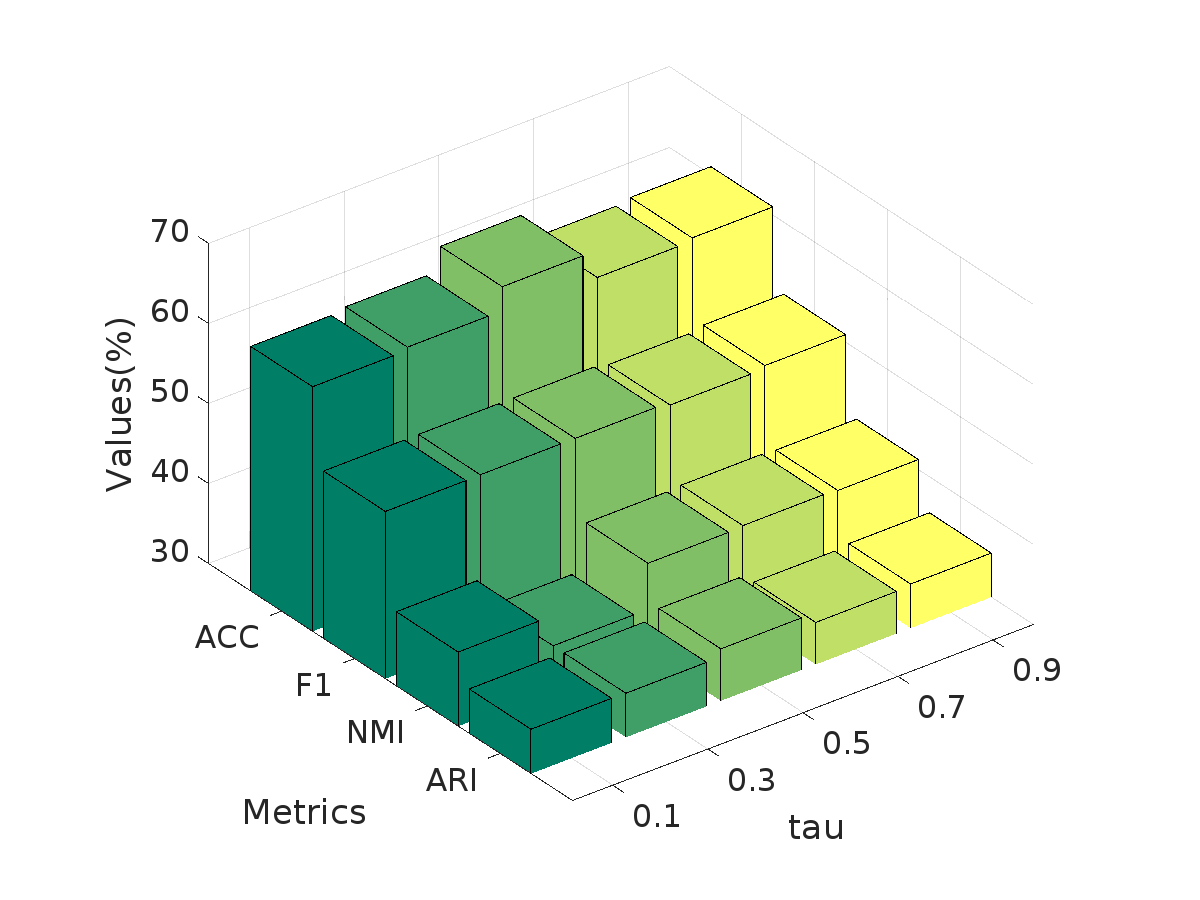}}
\centerline{(c) Wisconsin}
\end{minipage}
\begin{minipage}{0.49\linewidth}
\centerline{\includegraphics[width=0.98\textwidth]{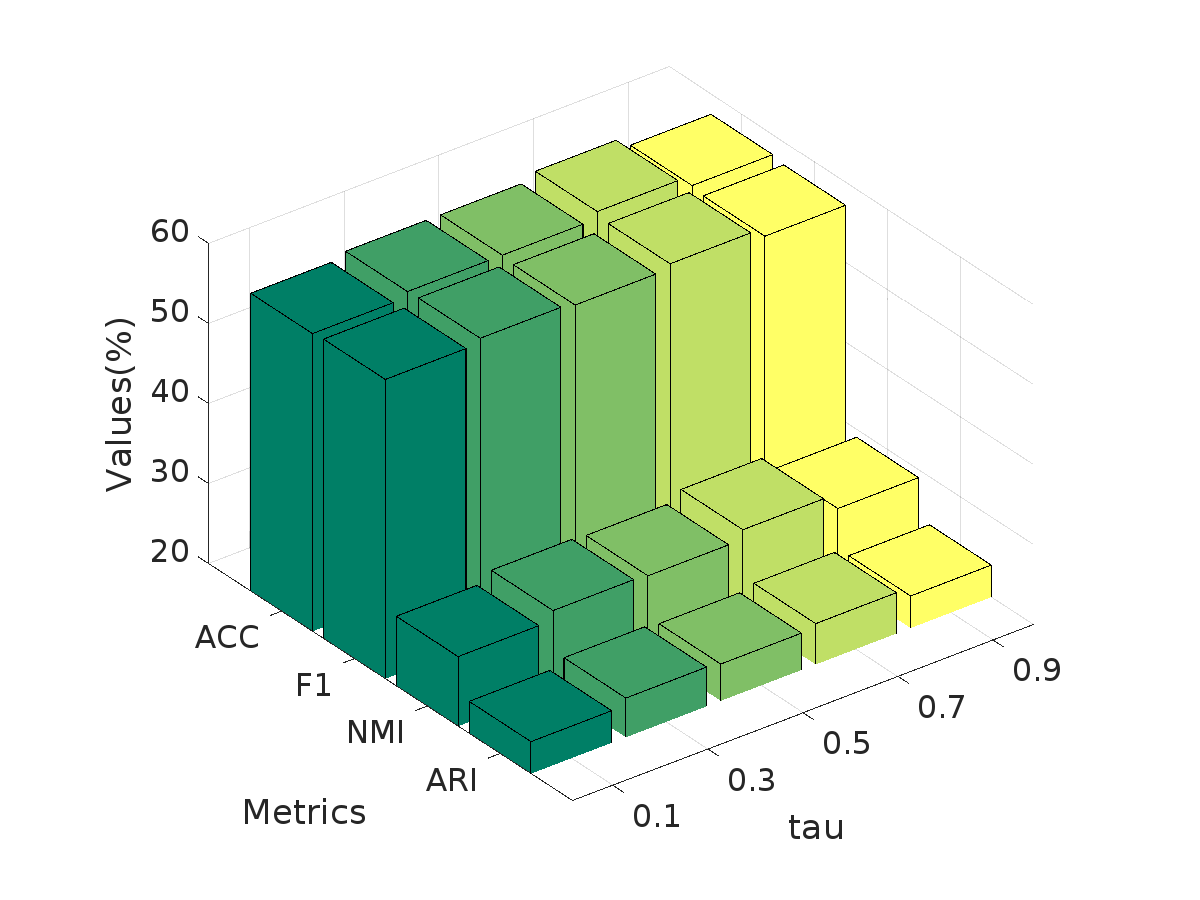}}
\centerline{(b) UAT}
\centerline{\includegraphics[width=0.98\textwidth]{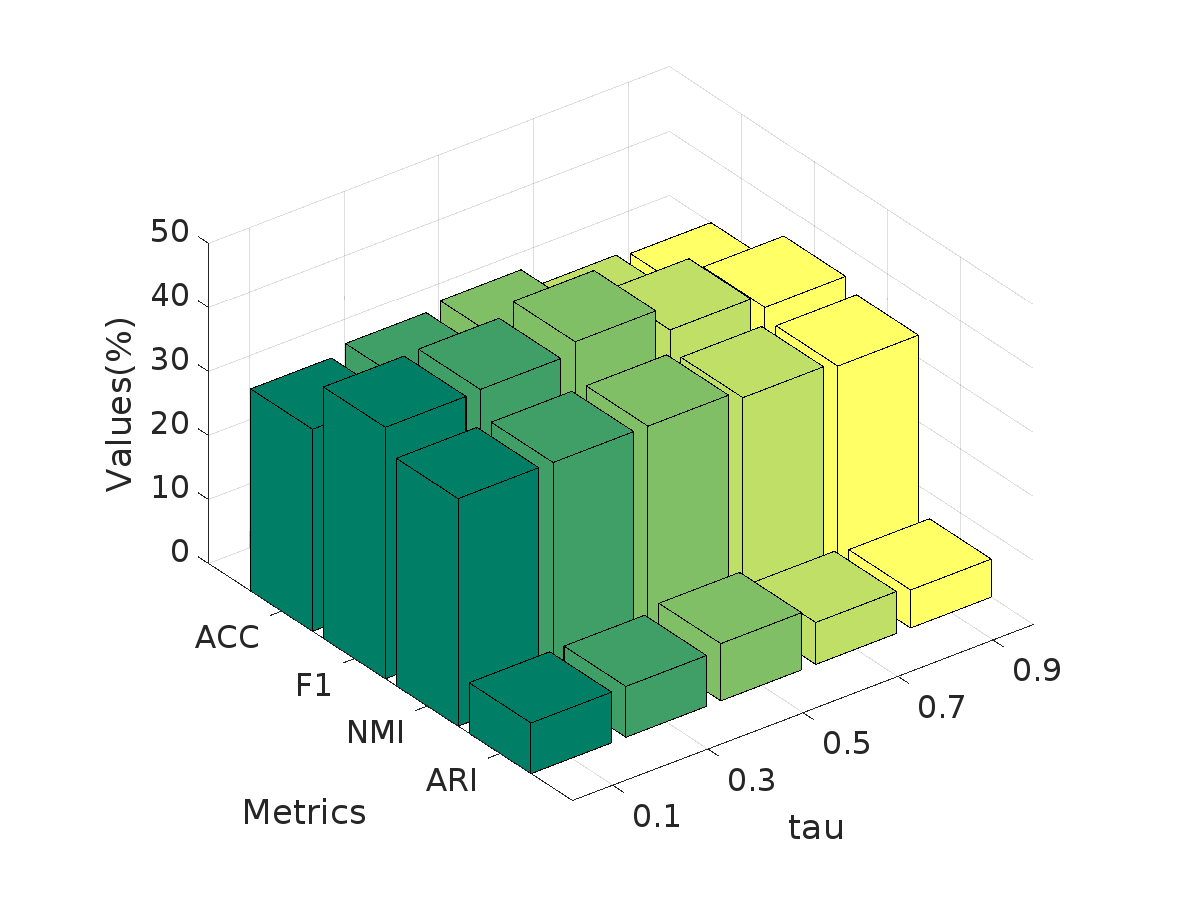}}
\centerline{(d) Squirrel}
\end{minipage}
\caption{Analysis of the confidence hyper-parameter $\tau$.}
\label{Sensitivity_tau}
\end{figure}

In this section, we analyze the hyper-parameters $\tau$ and $\beta$ in our method. The analysis of $\beta$ is shown in Figure~\ref{beta_an}. Our method achieves the best performance with $\beta = 0.3$ on the UAT dataset and $\beta = 0.2$ on the CITE dataset. Additionally, we conduct a hyper-parameter analysis for $\tau$, with the experimental results presented in Figure~\ref{Sensitivity_tau}. For the confidence $\tau$, we select it in $\{0.1, 0.3, 0.5, 0.7, 0.9\}$. As shown in figures, our method achieves promising performance when $\tau \in [0.7, 0.9]$ on the CORA dataset and when $\tau \in [0.5, 0.7]$ on the UAT dataset. For the Wisconsin and Squirrel datasets, the optimal performance is observed when $\tau \in [0.3, 0.5]$.
Besides, we observe that DCGC is not sensitive to the hyperparameter $\lambda$. Therefore, following previous work~\cite{liu2022deep}, we set $\lambda$ to 10.
In this paper, the confidence is set to a fixed value, thus a possible future work is to design a learnable or dynamical confidence. 

\subsection{Complexity Analysis}
In this section, we analyze the time and space complexity of the proposed dual-center graph clustering method. Here, we denote the batch size is $B$, the number of clusters is $K$, and the dimension of embeddings is $d$. 
The time complexity of representation learning and dual-center optimization is $\mathcal{O}(B^2d)$ and $\mathcal{O}(BKd)$, respectively. Therefore, the overall time complexity is $\mathcal{O}(B^2d + BKd)$.
Since $K<B$, the time consumption of dual-center optimization is significantly lower than that of representation learning. Besides, the space complexity of our proposed method is $\mathcal{O}(B^2)$. Thus, the proposed method will not bring the high time or space costs compared with the classical infoNCE loss. 
The detailed process of our proposed DCGC is summarized in Algorithm~\ref{ALGORITHM}.

\subsection{Visualization Analysis}
To further demonstrate the superiority of DCGC, we conduct a 2D t-SNE~\cite{van2008visualizing} analysis on the learned node embeddings, as shown in Figure \ref{t_SNE}. The results reveal that DCGC does not significantly increase intra-cluster density or inter-cluster distance due to the introduction of the neighbor distribution center. Instead, dual-center optimization improves clustering performance by slightly increasing intra-cluster distance, as the dual-center space enhances the model's accuracy in classifying boundary points across clusters.

\section{Conclusion}
In this paper, we propose a novel dual-center graph clustering approach with neighbor distribution.
% , which includes representation learning with neighbor distribution and dual-center optimization. 
Specifically, we utilize neighbor distribution as a supervision signal to mine hard negative samples in contrastive learning, which is reliable and enhances the effectiveness of representation learning. Furthermore, the neighbor distribution center is introduced alongside the feature center to jointly construct a dual-target distribution for dual-center optimization and fine-tune representation learning, which utilizes the intra-class consistency and reliability of neighbor distribution to improve the single-target distribution. Experiments demonstrate the effectiveness and superiority of the proposed method. Future research will explore the application of neighbor distribution to large-scale datasets.

%%%%%%%%%%%%%%%%%%%%%%%%%%%%%%%%%%%%%%%%%%%%%%%%%%%%%%%%%%%%%%%%%%%%%%%%

%%% Use this environment to include acknowledgements (optional).
%%% This will be omitted in doubleblind mode.

\begin{ack}
This research is supported by the National Natural Science Foundation of China under Grants 62172261 and the National Natural Science Foundation of China under Grant 62206267.
\end{ack}

%%%%%%%%%%%%%%%%%%%%%%%%%%%%%%%%%%%%%%%%%%%%%%%%%%%%%%%%%%%%%%%%%%%%%%%%

%%% Use this command to include your bibliography file.

\bibliography{mybibfile}

\begin{thebibliography}{43}
\providecommand{\natexlab}[1]{#1}
\providecommand{\url}[1]{\texttt{#1}}
\expandafter\ifx\csname urlstyle\endcsname\relax
  \providecommand{\doi}[1]{doi: #1}\else
  \providecommand{\doi}{doi: \begingroup \urlstyle{rm}\Url}\fi

\bibitem[Bo et~al.(2020)Bo, Wang, Shi, Zhu, Lu, and Cui]{bo2020structural}
D.~Bo, X.~Wang, C.~Shi, M.~Zhu, E.~Lu, and P.~Cui.
\newblock Structural deep clustering network.
\newblock In \emph{Proceedings of the Web Conference}, pages 1400--1410, 2020.

\bibitem[Chen et~al.(2020)Chen, Kornblith, Norouzi, and Hinton]{chen2020simple}
T.~Chen, S.~Kornblith, M.~Norouzi, and G.~E. Hinton.
\newblock A simple framework for contrastive learning of visual representations.
\newblock In \emph{Proceedings of the 37th International Conference on Machine Learning}, volume 119, pages 1597--1607, 2020.

\bibitem[Cui et~al.(2020)Cui, Zhou, Yang, and Liu]{cui2020adaptive}
G.~Cui, J.~Zhou, C.~Yang, and Z.~Liu.
\newblock Adaptive graph encoder for attributed graph embedding.
\newblock In \emph{Proceedings of the 26th International Conference on Knowledge Discovery and Data Mining}, pages 976--985, 2020.

\bibitem[Dong et~al.(2021)Dong, Ding, Jalaian, Ji, and Li]{dong2021adagnn}
Y.~Dong, K.~Ding, B.~Jalaian, S.~Ji, and J.~Li.
\newblock Adagnn: Graph neural networks with adaptive frequency response filter.
\newblock In \emph{Proceedings of the 30th International Conference on Information and Knowledge Management}, pages 392--401, 2021.

\bibitem[Gao et~al.(2021)Gao, Yao, and Chen]{gao2021simcse}
T.~Gao, X.~Yao, and D.~Chen.
\newblock Simcse: Simple contrastive learning of sentence embeddings.
\newblock \emph{arXiv preprint arXiv:2104.08821}, 2021.

\bibitem[Ghasedi~Dizaji et~al.(2017)Ghasedi~Dizaji, Herandi, Deng, Cai, and Huang]{ghasedi2017deep}
K.~Ghasedi~Dizaji, A.~Herandi, C.~Deng, W.~Cai, and H.~Huang.
\newblock Deep clustering via joint convolutional autoencoder embedding and relative entropy minimization.
\newblock In \emph{Proceedings of the International Conference on Computer Vision}, pages 5736--5745, 2017.

\bibitem[Gong et~al.(2022)Gong, Zhou, Tu, and Liu]{gong2022attributed}
L.~Gong, S.~Zhou, W.~Tu, and X.~Liu.
\newblock Attributed graph clustering with dual redundancy reduction.
\newblock In \emph{Proceedings of the 31st International Joint Conference on Artificial Intelligence}, pages 3015--3021, 2022.

\bibitem[Grover and Leskovec(2016)]{grover2016node2vec}
A.~Grover and J.~Leskovec.
\newblock node2vec: Scalable feature learning for networks.
\newblock In \emph{Proceedings of the 22nd International Conference on Knowledge Discovery and Data Mining}, pages 855--864, 2016.

\bibitem[Hartigan and Wong(1979)]{hartigan1979algorithm}
J.~A. Hartigan and M.~A. Wong.
\newblock Algorithm as 136: A k-means clustering algorithm.
\newblock \emph{Journal of the Royal Statistical Society. Series C (Applied Statistics)}, 28\penalty0 (1):\penalty0 100--108, 1979.

\bibitem[Kipf and Welling(2016{\natexlab{a}})]{kipf2016semi}
T.~N. Kipf and M.~Welling.
\newblock Semi-supervised classification with graph convolutional networks.
\newblock \emph{arXiv preprint arXiv:1609.02907}, 2016{\natexlab{a}}.

\bibitem[Kipf and Welling(2016{\natexlab{b}})]{kipf2016variational}
T.~N. Kipf and M.~Welling.
\newblock Variational graph auto-encoders.
\newblock \emph{arXiv preprint arXiv:1611.07308}, 2016{\natexlab{b}}.

\bibitem[Lee et~al.(2022)Lee, Lee, and Park]{lee2022augmentation}
N.~Lee, J.~Lee, and C.~Park.
\newblock Augmentation-free self-supervised learning on graphs.
\newblock In \emph{Proceedings of the 34th AAAI Conference on Artificial Intelligence}, volume~36, pages 7372--7380, 2022.

\bibitem[Liu et~al.(2022)Liu, Tu, Zhou, Liu, Song, Yang, and Zhu]{liu2022deep}
Y.~Liu, W.~Tu, S.~Zhou, X.~Liu, L.~Song, X.~Yang, and E.~Zhu.
\newblock Deep graph clustering via dual correlation reduction.
\newblock In \emph{Proceedings of the 36th AAAI Conference on Artificial Intelligence}, volume~36, pages 7603--7611, 2022.

\bibitem[Liu et~al.(2023)Liu, Yang, Zhou, Liu, Wang, Liang, Tu, Li, Duan, and Chen]{liu2023hard}
Y.~Liu, X.~Yang, S.~Zhou, X.~Liu, Z.~Wang, K.~Liang, W.~Tu, L.~Li, J.~Duan, and C.~Chen.
\newblock Hard sample aware network for contrastive deep graph clustering.
\newblock In \emph{Proceedings of the 35th AAAI Conference on Artificial Intelligence}, volume~37, pages 8914--8922, 2023.

\bibitem[Luan et~al.(2022)Luan, Hua, Lu, Zhu, Zhao, Zhang, Chang, and Precup]{luan2022revisiting}
S.~Luan, C.~Hua, Q.~Lu, J.~Zhu, M.~Zhao, S.~Zhang, X.~Chang, and D.~Precup.
\newblock Revisiting heterophily for graph neural networks.
\newblock In \emph{Proceedings of the 35th International Conference on Neural Information Processing Systems}, volume~35, pages 1362--1375, 2022.

\bibitem[Ma et~al.(2021)Ma, Liu, Zhao, Liu, Tang, and Shah]{ma2021unified}
Y.~Ma, X.~Liu, T.~Zhao, Y.~Liu, J.~Tang, and N.~Shah.
\newblock A unified view on graph neural networks as graph signal denoising.
\newblock In \emph{Proceedings of the 30th International Conference on Information and Knowledge Management}, pages 1202--1211, 2021.

\bibitem[Oord et~al.(2018)Oord, Li, and Vinyals]{oord2018representation}
A.~v.~d. Oord, Y.~Li, and O.~Vinyals.
\newblock Representation learning with contrastive predictive coding.
\newblock \emph{arXiv preprint arXiv:1807.03748}, 2018.

\bibitem[Pan et~al.(2018)Pan, Hu, Long, Jiang, Yao, and Zhang]{pan2018adversarially}
S.~Pan, R.~Hu, G.~Long, J.~Jiang, L.~Yao, and C.~Zhang.
\newblock Adversarially regularized graph autoencoder for graph embedding.
\newblock \emph{arXiv preprint arXiv:1802.04407}, 2018.

\bibitem[Pan et~al.(2019)Pan, Hu, Fung, Long, Jiang, and Zhang]{pan2019learning}
S.~Pan, R.~Hu, S.-f. Fung, G.~Long, J.~Jiang, and C.~Zhang.
\newblock Learning graph embedding with adversarial training methods.
\newblock \emph{IEEE transactions on cybernetics}, 50\penalty0 (6):\penalty0 2475--2487, 2019.

\bibitem[Pei et~al.(2020)Pei, Wei, Chang, Lei, and Yang]{pei2020geom}
H.~Pei, B.~Wei, K.~C.-C. Chang, Y.~Lei, and B.~Yang.
\newblock Geom-gcn: Geometric graph convolutional networks.
\newblock \emph{arXiv preprint arXiv:2002.05287}, 2020.

\bibitem[Perozzi et~al.(2014)Perozzi, Al-Rfou, and Skiena]{perozzi2014deepwalk}
B.~Perozzi, R.~Al-Rfou, and S.~Skiena.
\newblock Deepwalk: Online learning of social representations.
\newblock In \emph{Proceedings of the 20th International Conference on Knowledge Discovery and Data Mining}, pages 701--710, 2014.

\bibitem[Ribeiro et~al.(2017)Ribeiro, Saverese, and Figueiredo]{ribeiro2017struc2vec}
L.~F.~R. Ribeiro, P.~H.~P. Saverese, and D.~R. Figueiredo.
\newblock \emph{struc2vec}: Learning node representations from structural identity.
\newblock In \emph{Proceedings of the 23rd {ACM} {SIGKDD} International Conference on Knowledge Discovery and Data Mining}, pages 385--394, 2017.

\bibitem[Rozemberczki et~al.(2021)Rozemberczki, Allen, and Sarkar]{rozemberczki2021multi}
B.~Rozemberczki, C.~Allen, and R.~Sarkar.
\newblock Multi-scale attributed node embedding.
\newblock \emph{Journal of Complex Networks}, 9\penalty0 (2):\penalty0 cnab014, 2021.

\bibitem[Sen et~al.(2008)Sen, Namata, Bilgic, Getoor, Galligher, and Eliassi-Rad]{sen2008collective}
P.~Sen, G.~Namata, M.~Bilgic, L.~Getoor, B.~Galligher, and T.~Eliassi-Rad.
\newblock Collective classification in network data.
\newblock \emph{AI magazine}, 29\penalty0 (3):\penalty0 93--93, 2008.

\bibitem[Shchur et~al.(2018)Shchur, Mumme, Bojchevski, and G{\"u}nnemann]{shchur2018pitfalls}
O.~Shchur, M.~Mumme, A.~Bojchevski, and S.~G{\"u}nnemann.
\newblock Pitfalls of graph neural network evaluation.
\newblock In \emph{NeurIPS Workshop}, 2018.

\bibitem[Shen et~al.(2023)Shen, Sun, Pan, Zhou, and Yang]{shen2023neighbor}
X.~Shen, D.~Sun, S.~Pan, X.~Zhou, and L.~T. Yang.
\newblock Neighbor contrastive learning on learnable graph augmentation.
\newblock In \emph{Proceedings of the AAAI conference on artificial intelligence}, volume~37, pages 9782--9791, 2023.

\bibitem[Tian et~al.(2014)Tian, Gao, Cui, Chen, and Liu]{tian2014learning}
F.~Tian, B.~Gao, Q.~Cui, E.~Chen, and T.-Y. Liu.
\newblock Learning deep representations for graph clustering.
\newblock In \emph{Proceedings of the 28th AAAI Conference on Artificial Intelligence}, volume~28, pages 101--105, 2014.

\bibitem[Tu et~al.(2021)Tu, Zhou, Liu, Guo, Cai, Zhu, and Cheng]{tu2021deep}
W.~Tu, S.~Zhou, X.~Liu, X.~Guo, Z.~Cai, E.~Zhu, and J.~Cheng.
\newblock Deep fusion clustering network.
\newblock In \emph{Proceedings of the 33rd AAAI Conference on Artificial Intelligence}, volume~35, pages 9978--9987, 2021.

\bibitem[Van~der Maaten and Hinton(2008)]{van2008visualizing}
L.~Van~der Maaten and G.~Hinton.
\newblock Visualizing data using t-sne.
\newblock \emph{Journal of Machine Learning Research}, 9\penalty0 (11), 2008.

\bibitem[Velickovic et~al.(2019)Velickovic, Fedus, Hamilton, Li{\`{o}}, Bengio, and Hjelm]{velickovic2019deep}
P.~Velickovic, W.~Fedus, W.~L. Hamilton, P.~Li{\`{o}}, Y.~Bengio, and R.~D. Hjelm.
\newblock Deep graph infomax.
\newblock In \emph{Proceedings of the 7th International Conference on Learning Representations}, 2019.

\bibitem[Wang et~al.(2019)Wang, Pan, Hu, Long, Jiang, and Zhang]{wang2019attributed}
C.~Wang, S.~Pan, R.~Hu, G.~Long, J.~Jiang, and C.~Zhang.
\newblock Attributed graph clustering: A deep attentional embedding approach.
\newblock \emph{arXiv preprint arXiv:1906.06532}, 2019.

\bibitem[Wu et~al.(2021)Wu, Lin, Gao, Tan, Li, et~al.]{wu2021graphmixup}
L.~Wu, H.~Lin, Z.~Gao, C.~Tan, S.~Li, et~al.
\newblock Graphmixup: Improving class-imbalanced node classification on graphs by self-supervised context prediction.
\newblock \emph{arXiv preprint arXiv:2106.11133}, 2021.

\bibitem[Xia et~al.(2021)Xia, Wu, Wang, Chen, and Li]{xia2021progcl}
J.~Xia, L.~Wu, G.~Wang, J.~Chen, and S.~Z. Li.
\newblock Progcl: Rethinking hard negative mining in graph contrastive learning.
\newblock \emph{arXiv preprint arXiv:2110.02027}, 2021.

\bibitem[Xie et~al.(2016)Xie, Girshick, and Farhadi]{xie2016unsupervised}
J.~Xie, R.~Girshick, and A.~Farhadi.
\newblock Unsupervised deep embedding for clustering analysis.
\newblock In \emph{Proceedings of the 33nd International Conference on Machine Learning}, pages 478--487, 2016.

\bibitem[Yang et~al.(2021)Yang, Li, Liu, Niu, Wang, Cao, and Guo]{yang2021diverse}
L.~Yang, M.~Li, L.~Liu, B.~Niu, C.~Wang, X.~Cao, and Y.~Guo.
\newblock Diverse message passing for attribute with heterophily.
\newblock In \emph{Proceedings of the 35th International Conference on Neural Information Processing Systems}, pages 4751--4763, 2021.

\bibitem[Yang et~al.(2022)Yang, Wang, Liu, Wen, Meng, Zhou, Liu, and Zhu]{yang2022mixed}
X.~Yang, Y.~Wang, Y.~Liu, Y.~Wen, L.~Meng, S.~Zhou, X.~Liu, and E.~Zhu.
\newblock Mixed graph contrastive network for semi-supervised node classification.
\newblock \emph{ACM Transactions on Knowledge Discovery from Data}, 2022.

\bibitem[Yang et~al.(2023)Yang, Liu, Zhou, Wang, Tu, Zheng, Liu, Fang, and Zhu]{yang2023cluster}
X.~Yang, Y.~Liu, S.~Zhou, S.~Wang, W.~Tu, Q.~Zheng, X.~Liu, L.~Fang, and E.~Zhu.
\newblock Cluster-guided contrastive graph clustering network.
\newblock In \emph{Proceedings of the 35th AAAI Conference on Artificial Intelligence}, volume~37, pages 10834--10842, 2023.

\bibitem[Yang et~al.(2024)Yang, Min, Liang, Liu, Wang, Zhou, Wu, Liu, and Zhu]{yang2024graphlearner}
X.~Yang, E.~Min, K.~Liang, Y.~Liu, S.~Wang, S.~Zhou, H.~Wu, X.~Liu, and E.~Zhu.
\newblock Graphlearner: Graph node clustering with fully learnable augmentation.
\newblock In \emph{Proceedings of the 32nd ACM International Conference on Multimedia}, pages 5517--5526, 2024.

\bibitem[You et~al.(2020)You, Chen, Sui, Chen, Wang, and Shen]{you2020graph}
Y.~You, T.~Chen, Y.~Sui, T.~Chen, Z.~Wang, and Y.~Shen.
\newblock Graph contrastive learning with augmentations.
\newblock In \emph{Proceedings of the 33rd International Conference on Neural Information Processing Systems}, volume~33, pages 5812--5823, 2020.

\bibitem[Zhang et~al.(2022)Zhang, Li, Wang, Liu, Liu, Liu, and Zhu]{liliang_2}
J.~Zhang, L.~Li, S.~Wang, J.~Liu, Y.~Liu, X.~Liu, and E.~Zhu.
\newblock Multiple kernel clustering with dual noise minimization.
\newblock In \emph{Proceedings of the 30th ACM International Conference on Multimedia}, pages 3440--3450, 2022.

\bibitem[Zhao et~al.(2021)Zhao, Yang, Wang, Yang, and Deng]{zhao2021graph}
H.~Zhao, X.~Yang, Z.~Wang, E.~Yang, and C.~Deng.
\newblock Graph debiased contrastive learning with joint representation clustering.
\newblock In \emph{Proceedings of the 30th International Joint Conference on Artificial Intelligence}, pages 3434--3440, 2021.

\bibitem[Zhou et~al.(2020)Zhou, Liu, Li, Zhu, Liu, Zhang, and Yin]{ZHOU_1}
S.~Zhou, X.~Liu, M.~Li, E.~Zhu, L.~Liu, C.~Zhang, and J.~Yin.
\newblock Multiple kernel clustering with neighbor-kernel subspace segmentation.
\newblock \emph{IEEE Transactions on Neural Networks and Learning Systems}, 31\penalty0 (4):\penalty0 1351--1362, 2020.

\bibitem[Zhu et~al.(2020)Zhu, Xu, Yu, Liu, Wu, and Wang]{zhu2020deep}
Y.~Zhu, Y.~Xu, F.~Yu, Q.~Liu, S.~Wu, and L.~Wang.
\newblock Deep graph contrastive representation learning.
\newblock \emph{arXiv preprint arXiv:2006.04131}, 2020.

\end{thebibliography}

\end{document}